\documentclass{article}

\usepackage{hyperref,graphicx,amsmath,amssymb,amsthm,subcaption,booktabs,cleveref,xspace,xcolor,float}
\usepackage[preprint]{neurips_2026}

\newcommand{\Mod}[1]{\ (\mathrm{mod}\ #1)}
\newcommand{\out}{\mathrm{out}}
\newcommand{\ti}{^{(i)}}
\newcommand{\sign}{\operatorname{sign}}

\newcommand{\coscos}{Cos$\to$Cos\xspace}
\newcommand{\sqcos}{Sq$\to$Cos\xspace}
\newcommand{\sqsq}{Sq$\to$Sq\xspace}

\newcommand{\relu}{\mathrm{ReLU}}
\newcommand{\E}{\mathbb{E}}
\newcommand{\Var}{\operatorname{Var}}
\newcommand{\argmax}{\operatorname{argmax}}

\newtheorem{lemma}{Lemma}

\title{Latent Algorithmic Structure Precedes Grokking: A Mechanistic Study of
ReLU MLPs on Modular Arithmetic}

\author{%
    Anand~Swaroop \\
    \texttt{anandswaroop191@gmail.com}
}

\begin{document}

\maketitle

\begin{abstract}
Grokking on modular addition has been characterized by sinusoidal
representations in transformers and MLPs. We find that ReLU MLPs in our
setting instead learn near-binary square-wave input weights, with
intermediate values appearing only near sign-change boundaries, and output
weights whose dominant Fourier phases satisfy $\phi_{\out} \approx \phi_a +
\phi_b$. These relations persist under label noise and even in models that
do not grok. Using DFT-extracted frequencies and phases to replace learned
weights with ideal square/cosine waves yields \textbf{95.8\% accuracy} from
noisy models that achieve only 0.40\%. This suggests that grokking sharpens
an algorithm already encoded during memorization.
\end{abstract}

\section{Introduction}

Neural networks trained on modular arithmetic tasks exhibit a counterintuitive
phenomenon: After rapidly memorizing the training set, validation accuracy
remains near random chance for thousands of additional steps before abruptly
rising to near-perfect generalization \citep{power2022}. This delayed
generalization, termed \emph{grokking}, has become a testbed for understanding
how and when neural networks develop structured internal representations.

Mechanistic interpretability work on transformers has characterized one
algorithm underlying grokking in some models: The network learns embeddings that
are sinusoids and computes the sum via trigonometric identities
\citep{nanda2023}. This algorithm has been shown to arise in non-neural models
and those trained without gradient descent \citep{mallinar2025}. This picture
was extended to MLPs by \citet{gromov2023}, which showed that sinusoidal weights
are sufficient for perfect accuracy modular addition in an MLP with a
sufficiently large hidden layer, derived an analytic solution for quadratic
activations, and showed empirically that this extends to ReLU. Together, these
works establish sinusoidal weights as a common signature of grokking on modular
addition tasks. However, while \citet{gromov2023} demonstrate that trained
weights are periodic and match an analytic cosine solution in the quadratic
model, they do not systematically characterize the precise waveform of the
learned periodic weights (e.g. cosine vs square) in ReLU MLPs.

Prior work has shown that phases in MLPs that grok the modular arithmetic task
often follow phase relations, such as $\phi_a + \phi_b = \phi_{\out}$
\citep{gromov2023,he2026}. Grokking on modular arithmetic has often been
described as a ``lottery ticket'' mechanism, where each individual hidden
neuron's outputs are noisy, but the number of neurons cancels out the noise
\citep{he2026,minegishi2025}.

\citet{li2025} prove that Fourier circuits emerge in one-hidden-layer MLPs with
polynomial activations via margin maximization, establishing a theoretical basis
for the sinusoidal structure in one-hidden-layer MLPs on modular addition.
\citet{doshi2024} study transformers with quadratic activations, showing that
memorizing and generalizing neurons coexist and can be identified via their
Fourier spectra. While their work characterizes the \emph{presence} of periodic
structure under noise, it does not examine the weight shape, the phase-sum
constraint, or whether the algorithmic structure survives in models that fail to
grok entirely.

\textbf{Contribution.} We study ReLU MLPs on the same modular addition task
and find a qualitatively different weight structure. Input weights are
\emph{near-binary square waves} rather than cosines, with intermediate
values appearing exclusively near sign-change boundaries. For each hidden
neuron $i$, let $\phi_a\ti$, $\phi_b\ti$, and $\phi_\out\ti$ denote the
phases of the dominant Fourier components of the input weights over $a$, the
input weights over $b$, and the output weights respectively. We find that
$\phi_\out\ti = \phi_a\ti + \phi_b\ti$ holds for structured neurons across
all noise levels, including models that fail to generalize entirely.
Extracting these phases and frequencies via DFT and replacing all weights
with ideal waves recovers \textbf{95.8\% accuracy} from models achieving
only \textbf{0.40\%}, and substantial accuracy from saturation checkpoints
before grokking occurs. This demonstrates that the correct algorithm is
substantially encoded during memorization; grokking sharpens latent
structure rather than discovering it.

\section{Method}
\label{sec:method}

We train a multi-layer perceptron (MLP) with one hidden layer on the task of
adding two integers, $a$ and $b$, modulo some positive integer $p$. The input
is a dimension-$2p$ two-hot vector (a one-hot encoding of $a$ concatenated with
a one-hot encoding of $b$). The output is a dimension-$p$ one-hot vector
representing $a + b \Mod{p}$. The hidden layer consists of $N$ fully connected
neurons and uses the ReLU activation function.

For each hidden neuron $i$, we write $W_a\ti$ and $W_b\ti$ for the vectors of
input weights over the one-hot encodings of $a$ and $b$ respectively, and
$W_\out\ti$ for the vector of weights connecting neuron $i$ to each of the $p$
output nodes. We write $\phi_a\ti$, $\phi_b\ti$, $\phi_\out\ti$ for the phases
of the dominant Fourier components of $W_a\ti$, $W_b\ti$, $W_\out\ti$
respectively, extracted via discrete Fourier transform.

We train a one-hidden-layer ReLU MLP with AdamW (learning rate $10^{-3}$,
weight decay $1.0$, no scheduler, cross-entropy loss, full-batch training)
on modular addition modulo $p=97$ with $N=256$ hidden neurons. The dataset
is the full set of $97^2 = 9409$ pairs $(a,b)$, split 30/70 into train and
validation sets stratified by $c^* = a+b \Mod{97}$. We train a clean model
(noise level $\alpha=0$) and six noisy models
($\alpha\in\{0.05,0.10,0.15,0.20,0.25,0.30\}$), corrupting each selected
label uniformly among the other 96 classes. Training stops after 50{,}000
steps without a validation accuracy improvement of at least $10^{-4}$ or at
500{,}000 steps; we save a saturation checkpoint when the model first
achieves 99\% train accuracy, as well as the final checkpoint after training
ends. Results are consistent across 15 random seeds.

\section{Results}

\begin{figure}[h]
    \centering
    \includegraphics[width=\linewidth]{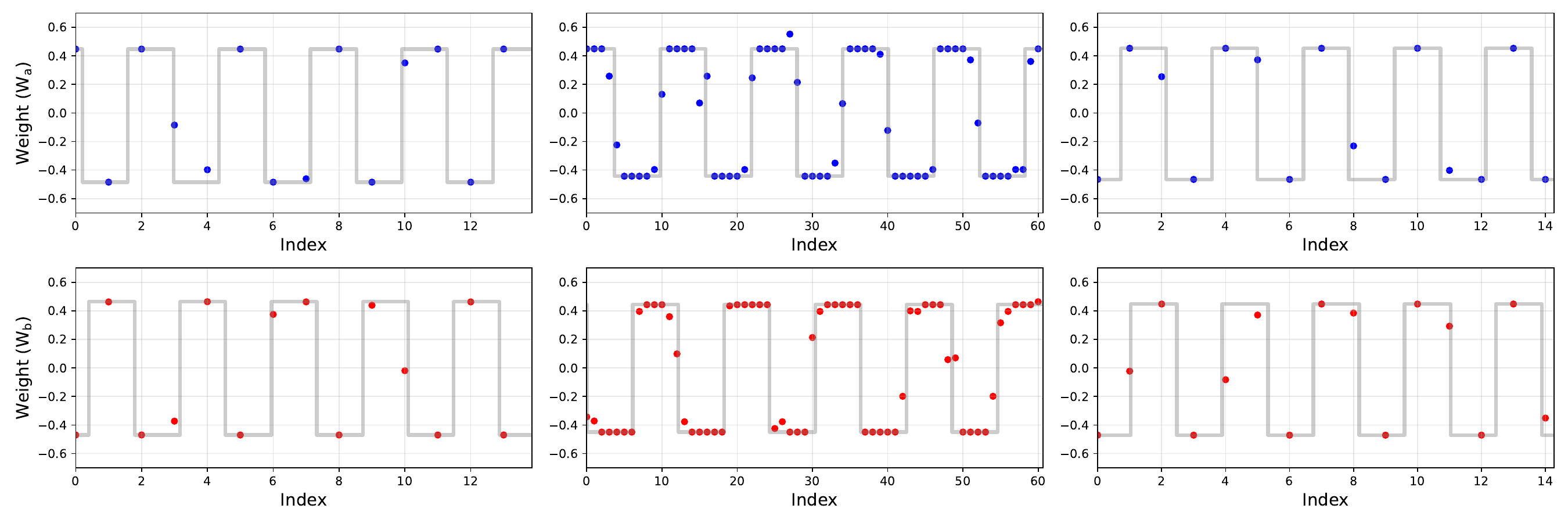}
    \caption{Actual $W_a$ for various neurons plotted in blue, actual $W_b$ in
        red, and ideal square wave in gray. Intermediate values appear only near
        sign-change boundaries. Figures are scaled horizontally such that no
        more than 5 periods are visible in order to show fine details.}
    \label{fig:wsquare}
\end{figure}

We find that in the clean model, $W_a\ti$ and $W_b\ti$ follow square waves
almost perfectly, as opposed to the sinusoidal waves found in the embeddings
of grokking transformers in \citet{nanda2023}. For each neuron, we extract
the frequency and phase of the dominant Fourier component with DFT and
construct a square wave with the same frequency and phase; no fitting is
performed on the frequency and phase. We compute the amplitude and vertical
offset such that the constructed wave's upper and lower levels match the
90-th percentile weights out of all positive weights and the 10-th
percentile weights out of all negative weights, respectively. We call these
constructed waves \textit{ideal square waves}. Several examples of real
neuron weights with overlaid ideal square waves are shown in
\Cref{fig:wsquare}.

Not all neurons in the hidden layer learn periodic representations. To
distinguish between neurons that encode useful frequency structure and those
that do not, we quantify the periodicity of each neuron's input weights using
the \emph{periodicity score}: The ratio of the magnitude of the dominant Fourier
component to the mean magnitude across all non-DC components,
\[
    \mathrm{per}(w) = \frac{\max_{k=1}^{p-1} |\hat{w}_k|}
    {\frac{1}{p-1}\sum_{k=1}^{p-1} |\hat{w}_k|},
\]
where $\hat{w}_k$ denotes the (complex-valued) $k$-th DFT coefficient of
weight vector $w$. A high periodicity score indicates that the weight vector
is dominated by a single frequency; a lower score indicates a flatter
spectrum with no particularly dominant frequency. Random values tend to
score 2--3.

Applying this measure to $W_a\ti$ reveals a bimodal distribution: most
neurons have high periodicity scores ($\mathrm{per}(W_a\ti)>7.5$) and the
remainder have low scores ($\mathrm{per}(W_a\ti)<5$), with the same split on
$W_b\ti$. We call the high-periodicity neurons \emph{structured} and the
rest \emph{unstructured}. Zeroing the outputs of unstructured neurons
changes accuracy by only 0.0004 on average across 15 seeds, and their output
weights have smaller norms (see \Cref{fig:periodicity_alpha0.00} in
Appendix~\ref{apx:periodicity}). Thresholds between 6 and 9 yield nearly
identical groupings and yield a change in accuracy of no more than $0.001$,
so we use the threshold $7.5$ throughout for simplicity (we discuss the
periodicity distribution in Appendix~\ref{apx:periodicity}).

We find unstructured neurons tend to have significantly lower-norm
connections to the output layer, as shown in
\Cref{fig:periodicity_alpha0.00} in Appendix~\ref{apx:periodicity}. Across
all 15 seeds, the dominant Fourier frequencies of $W_a\ti$, $W_b\ti$, and
$W_{\out}\ti$ are equal for $95.94\%$ of all neurons (standard deviation
$0.84\%$) and for 100\% of structured neurons, and the dominant frequencies
are roughly uniform on $[1, 48]$. This suggests some algorithmic structure
is present in the weights, which we explain further in \Cref{sec:ideal}.

To provide a quantitative measure of fit, for each neuron, we calculate the
mean squared deviation (MSD) between the input weights and the ideal square
wave, as well as the MSD between the input weights and an ideal cosine wave
with the same phase, frequency, and vertical offset, with amplitude
multiplied by a factor of $\frac{4}{\pi}$ (this is the amplitude that
minimizes the MSD of the ideal cosine wave against the ideal square wave; we
provide a proof of this in \Cref{lma:4pi} in Appendix~\ref{apx:distance}).
We denote the MSD difference as the square MSD minus the cosine MSD. We
perform a two-sample $t$-test of structured neurons against unstructured
neurons and find that the MSD difference lower in structured neurons across
all 15 seeds, with $p$-value less than $10^{-5}$ across all seeds with a
one-tailed test. We provide full numerical results and a more detailed
discussion of the metric in Appendix~\ref{apx:distance}. In addition, we
define the metric of \textit{bimodality} as the fraction of input weights
that lie within 10\% of the 90-th percentile weight in the positive region
(that is, in the range $[0.9P, 1.1P]$ where $P$ is the 90-th percentile
positive weight), or within 10\% of the 10-th percentile weight in the
negative region. This effectively counts the fraction of input weights that
are near the extremes while ignoring extreme outliers. Bimodality of
$W_a\ti$ is plotted against periodicity of $W_a\ti$ in
\Cref{fig:bimodality}: The two clusters of structured and unstructured
neurons are clear, and periodicity appears directly correlated to
bimodality. This further suggests that the weight shape present in the
structured neurons is a square wave.

\subsection{Noise}

\begin{figure}[h]
    \centering
    \begin{subfigure}{0.45\textwidth}
        \centering
        \includegraphics[width=\linewidth]{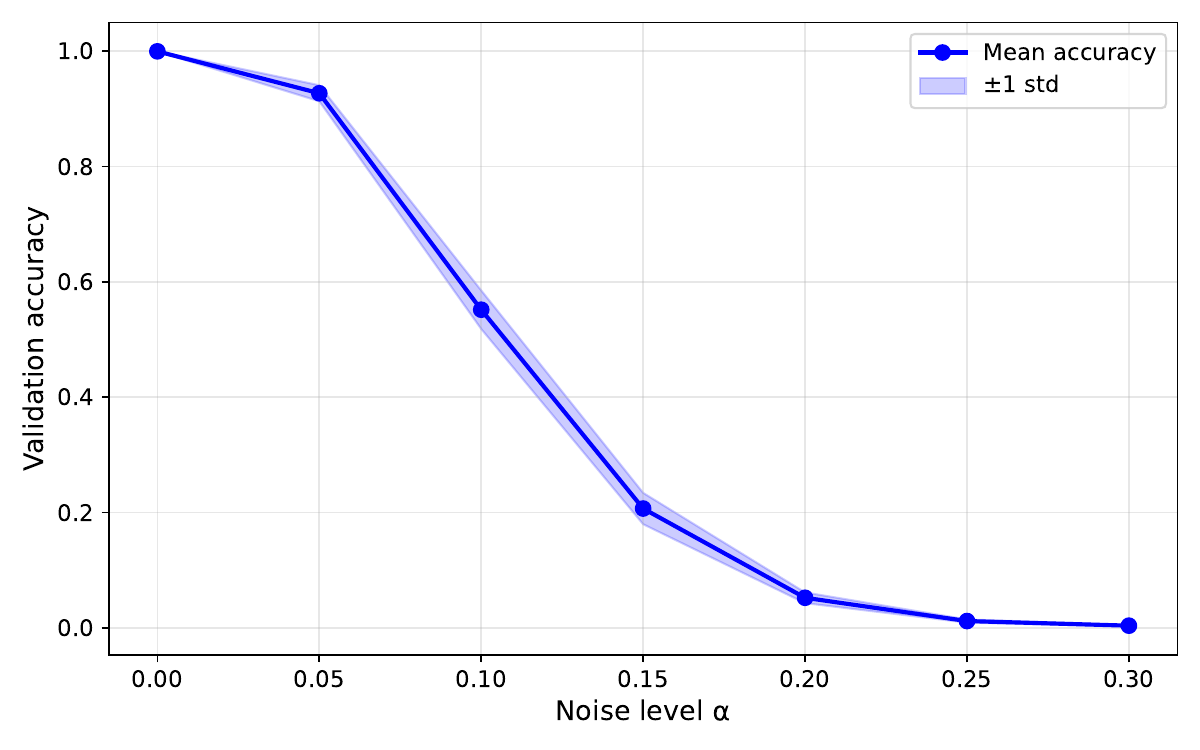}
        \caption{Validation accuracy against $\alpha$.}
        \label{fig:accnoise}
    \end{subfigure}
    \hfill
    \begin{subfigure}{0.45\textwidth}
        \centering
        \includegraphics[width=\linewidth]{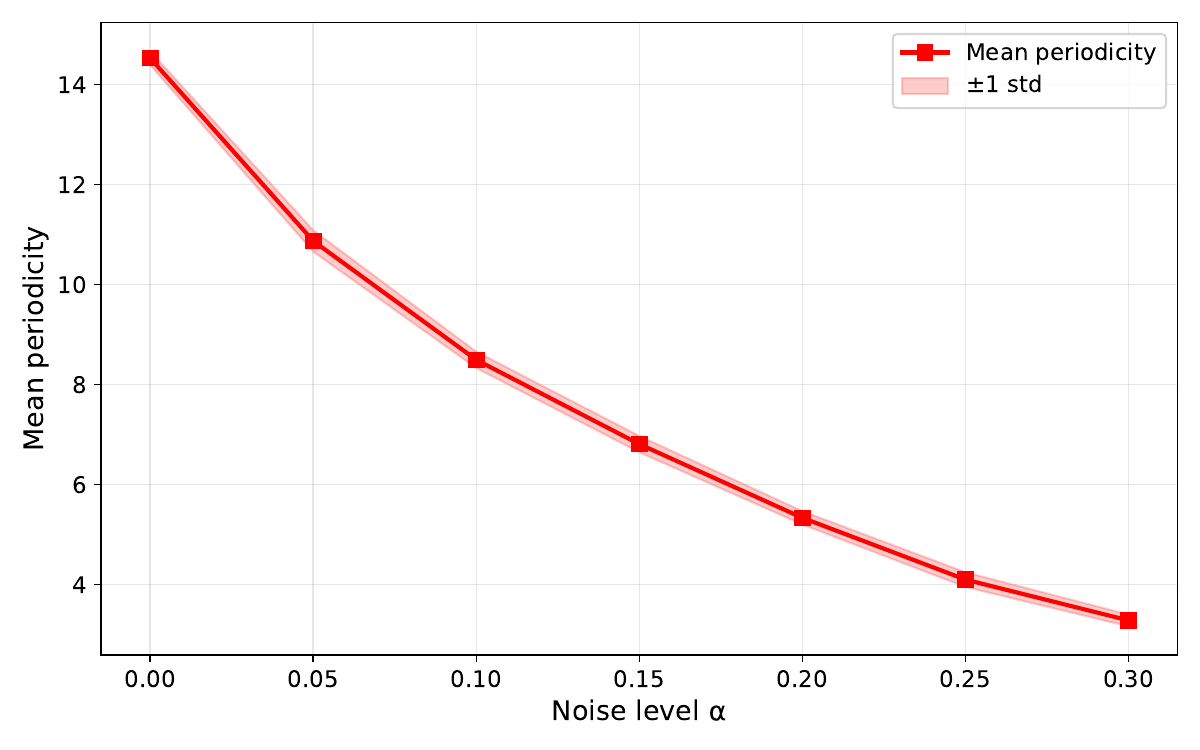}
        \caption{Mean periodicity over all neurons against $\alpha$.}
        \label{fig:periodicitynoise}
    \end{subfigure}
    \caption{Accuracy and periodicity across noise levels. Mean and standard
    deviation across 15 seeds are plotted.}
\end{figure}

We find that as the noise level $\alpha$ increases, the validation accuracy
of the final model follows a decreasing sigmoidal transition
(\cref{fig:accnoise}) from $>0.999$ with $\alpha = 0.00$ to $<0.01$ with
$\alpha = 0.30$, while periodicity of neurons follows a gradual decrease
(\cref{fig:periodicitynoise}).

\begin{figure}[t]
\centering
\begin{minipage}{0.58\linewidth}
    \centering
    \includegraphics[width=\linewidth]{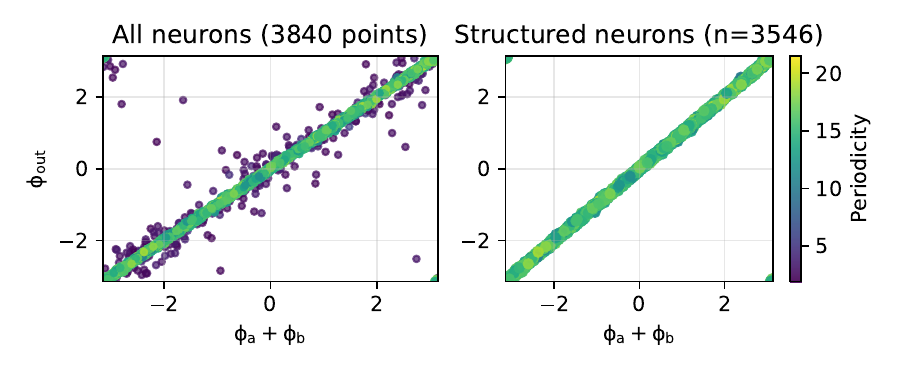}
    \caption{$\phi_{\out}$ against $\phi_a + \phi_b$ across all neurons in the final checkpoint of the
    clean model across all 15 seeds, wrapped to $[-\pi, \pi]$. When restricted
    to only structured neurons, there is a nearly perfect correlation (circular
    correlation $r = 0.9986$).}
    \label{fig:phase}
\end{minipage}
\hfill
\begin{minipage}{0.38\linewidth}
    \centering
    \includegraphics[width=\linewidth]{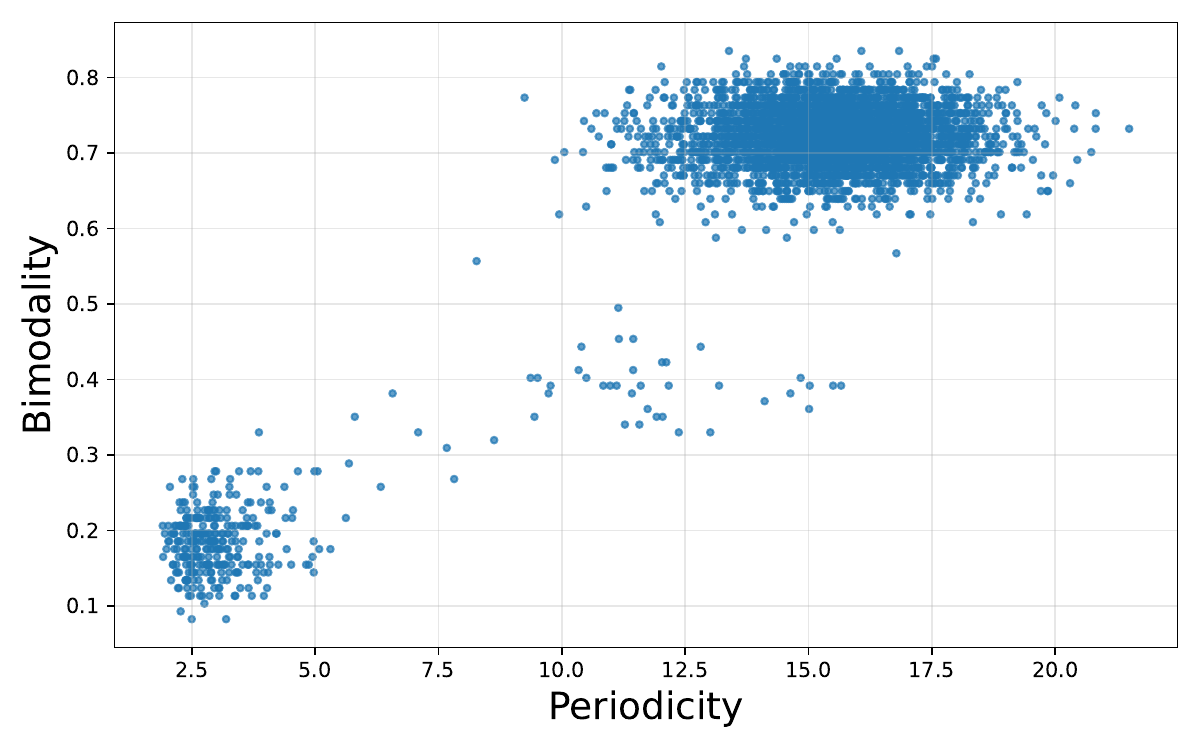}
    \caption{Bimodality against periodicity of $W_a\ti$ in the final checkpoint of the clean model (pooled across all neurons on all 15 seeds).}
    \label{fig:bimodality}
\end{minipage}
\end{figure}

\begin{figure}
    \centering
    \includegraphics[width=\linewidth]{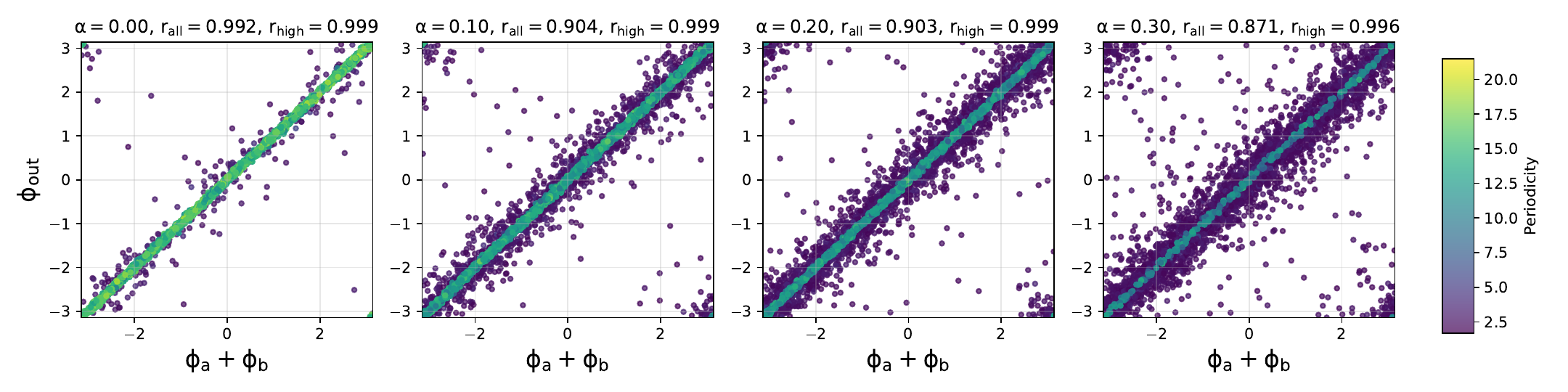}
    \caption{$\phi_{\out}$ against $\phi_a + \phi_b$ for various $\alpha$
    levels. Circular correlation coefficients are listed; $r_{\mathrm{all}}$ is taken
    over all neurons, $r_{\mathrm{high}}$ over only structured neurons
    (periodicity $> 7.5$). $r_{\mathrm{high}} \ge 0.996$ for all $\alpha$ levels,
    indicating that structured neurons satisfy the phase-sum relation regardless
    of noise.}
    \label{fig:phasenoise}
\end{figure}

We find that, across all noise levels, the dominant frequencies of $W_a\ti$,
$W_b\ti$, and $W_{\out}\ti$ match for 100\% of neurons with periodicity $>
7.5$, although the number of structured neurons decreases monotonically with
noise. We extract the phase of the dominant Fourier component from each
neuron's $W_{\out}$ (which, as explained previously, has the same frequency
as the dominant components of $W_a$ and $W_b$). We find that $\phi_{\out} =
\phi_a + \phi_b$ holds almost exactly for the highest periodicity neurons in
both the clean model and the noisy models. As shown in \Cref{fig:phase}, all
structured neurons satisfy $\phi_{\out} = \phi_a + \phi_b$---up to shifts by
multiples of $2\pi$---almost perfectly. This matches the empirical results
uncovered in \citet{gromov2023} and \citet{nanda2023} and the analytical
construction in the former. However, both works find this relation in
sinusoidal weights rather than square-wave-like weights.

Although periodicity generally degrades as the label noise increases, we
find that the $\phi_{\out} = \phi_a + \phi_b$ relation is preserved for the
highest-periodicity neurons within each tested $\alpha$ level. This is
evident in \Cref{fig:phasenoise}. This suggests that latent structure is
formed to some degree, even with significant label noise, which raises the
question of whether it can be extracted to recover a correct algorithm
despite the model's failure to generalize. We find that this is possible.

\section{Idealized Model}
\label{sec:ideal}

We first construct from scratch an MLP that solves the modular addition
task. We use an MLP with the same setup (dimension-$2p$ two-hot input,
dimension-$N$ hidden layer, dimension-$p$ one-hot output). We use weights
\begin{align*}
    W_a\ti[c] &= f\left(\frac{2 \pi i c}{p} - \phi_a\ti\right), & i \in [0, N - 1], c \in [0, p - 1] \\
    W_b\ti[c] &= f\left(\frac{2 \pi i c}{p} - \phi_b\ti\right), \\
    W_\out\ti[c] &= g\left(\frac{2 \pi i c}{p} - \phi_a\ti - \phi_b\ti\right), \\
\end{align*}
with phases $\phi_a\ti$ and $\phi_b\ti$ chosen independently per neuron and
uniformly at random from the range $[-\pi, \pi]$. We set all biases to zero: In
the trained clean model across all seeds, the effective first-layer bias (after
accounting for the vertical offset of the input weights) has mean $0.0092$ and
standard deviation $0.0050$ across structured neurons, less than 2.5\% of the
typical weight amplitude $A \approx 0.45$, and all structured neurons
satisfy $|b_{\mathrm{eff}}| < 0.06$.

As there are only 97 classes and 256 neurons, the frequencies of the
constructed weights alias; this construction is equivalent to picking
frequencies sequentially from $0, 1, \dots, 48$. We retain this somewhat
convoluted construction for consistency with \citet{gromov2023}. In fact,
the exact frequencies seem to have no effect on the final results beyond the
fact that they're uniformly distributed; when constructing models with
frequencies sampled uniformly at random from the range $[0, 48]$, we find
the same quantitative and qualitative results. Therefore, to avoid
redundancy, we present only the results for constructed models with
frequencies equal to the neuron index $i$.

$f$ and $g$ determine the type of wave that $W_a\ti$, $W_b\ti$,
$W_{\out}\ti$ follow: $\cos$ for sinusoidal waves and $\sign \circ \cos$ for
square waves. We test three different settings: $f = g = \cos$; $f = \sign
\circ \cos$, $g = \cos$; and $f = g = \sign \circ \cos$. We will refer to
these three as \coscos, \sqcos, and \sqsq respectively. The \coscos
construction is identical to that of \citet{gromov2023}. The goal of using
$\sign \circ \cos$ is to determine how square waves affect the model's
accuracy in comparison to sinusoidal waves.

We evaluate the constructed models over 40 different random seeds and
various hidden layer widths $N$, using both ReLU and quadratic activation
functions; the results are summarized in \Cref{tab:constructed_accuracy}.
All three methods show a monotonic increase in accuracy as the hidden layer
width increases, and all achieve high ($> 0.998$) accuracy with 512 neurons
for both activations. \sqcos shows the greatest accuracy across all hidden
widths with ReLU activations, and \coscos shows the greatest accuracy across
all hidden widths with quadratic activations.

This suggests that not only is the square wave construction for input weights
more faithful to the representation our model learns, but also that it is a more
accurate solution (for ReLU activations). This also suggests that ReLU
activations are necessary for square waves to appear: As square input waves
yield higher accuracy with ReLU, model input weights may tend to square waves
during training to minimize accuracy loss; with the quadratic activation
function, sinusoidal input weights yield higher accuracy, so training may push
the weights toward sinusoidal weights rather than square waves to minimize
accuracy loss, yielding the sinusoidal waves observed in past works.

\begin{table}[h]
\centering
\tiny
\caption{Accuracy of constructed models for different hidden widths. Values are mean $\pm$ standard deviation over 40 random seeds. Compares ReLU and quadratic activations.}
\label{tab:constructed_accuracy}
\begin{tabular}{c|ccc|ccc}
\toprule
& \multicolumn{3}{c|}{ReLU} & \multicolumn{3}{c}{Quadratic} \\
$N$ & \coscos & \sqcos & \sqsq & \coscos & \sqcos & \sqsq \\
\midrule
16 & $0.0422 \pm 0.0049$ & $\mathbf{0.0735 \pm 0.0041}$ & $0.0458 \pm 0.0064$ & $\mathbf{0.1954 \pm 0.0294}$ & $0.1589 \pm 0.0219$ & $0.1069 \pm 0.0216$ \\
32 & $0.1322 \pm 0.0125$ & $\mathbf{0.1944 \pm 0.0093}$ & $0.1415 \pm 0.0107$ & $\mathbf{0.6153 \pm 0.0514}$ & $0.4349 \pm 0.0479$ & $0.3178 \pm 0.0464$ \\
64 & $0.3653 \pm 0.0313$ & $\mathbf{0.4449 \pm 0.0234}$ & $0.3314 \pm 0.0178$ & $\mathbf{0.9562 \pm 0.0148}$ & $0.8015 \pm 0.0372$ & $0.6543 \pm 0.0518$ \\
96 & $0.5850 \pm 0.0379$ & $\mathbf{0.6522 \pm 0.0263}$ & $0.5057 \pm 0.0245$ & $\mathbf{0.9965 \pm 0.0021}$ & $0.9451 \pm 0.0163$ & $0.8460 \pm 0.0403$ \\
128 & $0.7482 \pm 0.0372$ & $\mathbf{0.7968 \pm 0.0243}$ & $0.6512 \pm 0.0258$ & $\mathbf{0.9997 \pm 0.0004}$ & $0.9863 \pm 0.0078$ & $0.9411 \pm 0.0207$ \\
192 & $0.9222 \pm 0.0178$ & $\mathbf{0.9417 \pm 0.0100}$ & $0.8450 \pm 0.0196$ & $\mathbf{1.0000 \pm 0.0000}$ & $0.9995 \pm 0.0006$ & $0.9922 \pm 0.0061$ \\
256 & $0.9793 \pm 0.0067$ & $\mathbf{0.9849 \pm 0.0042}$ & $0.9375 \pm 0.0107$ & $\mathbf{1.0000 \pm 0.0000}$ & $1.0000 \pm 0.0001$ & $0.9993 \pm 0.0007$ \\
384 & $0.9987 \pm 0.0008$ & $\mathbf{0.9992 \pm 0.0005}$ & $0.9907 \pm 0.0030$ & $\mathbf{1.0000 \pm 0.0000}$ & $1.0000 \pm 0.0000$ & $1.0000 \pm 0.0000$ \\
512 & $0.9999 \pm 0.0001$ & $\mathbf{1.0000 \pm 0.0001}$ & $0.9989 \pm 0.0006$ & $\mathbf{1.0000 \pm 0.0000}$ & $1.0000 \pm 0.0000$ & $1.0000 \pm 0.0000$ \\
\bottomrule
\end{tabular}
\end{table}

\subsection{Theoretical Analysis}
\label{sec:theory}

To justify these theoretical constructions with ReLU activations, we prove
that, as $N$ approaches $\infty$, the accuracy of each construction
(\coscos, \sqcos, \sqsq) becomes perfect almost surely. The key idea is to
compute the expected single-neuron contribution, $\E[X_c\ti]$, to each
output class $c$, and show that as $N$ approaches $\infty$, the sum of
correct-class contributions $\sum_i \E[X_{c^*}\ti]$ grows faster than
$\sum_i \E[X_c\ti]$ for any $c \ne c^*$ (full proofs in
Appendix~\ref{apx:theory}).

Additionally, we compute $\E[X_{c^*}\ti]$ and $\Var(X_{c^*}\ti)$ in order to
calculate the signal-to-noise ratio (SNR), equal to
$\frac{\E[X_{c^*}\ti]}{\sqrt{\Var(X_{c^*}\ti)}}$ (\cref{tab:snr}). This
provides a heuristic to predict the performance of each construction at
finite width. We find that the SNRs follow the hierarchy \sqcos $>$ \coscos
$>$ \sqsq, which aligns with the accuracy ordering we observe with
each construction in practice.

\begin{table}[h]
\centering
\scriptsize
\caption{Theoretical expected value, variance, and SNR for each neuron's
contribution to the correct class logit.}
\label{tab:snr}
\begin{tabular}{lccc}
\toprule
Construction & Expected value & Variance & SNR \\
\midrule
\coscos & $\approx 0.1351$ & $\approx 0.2317$ & $\approx 0.2806$ \\
\sqcos  & $\approx 0.2026$ & $\approx 0.4589$ & $\mathbf{\approx 0.2991}$ \\
\sqsq   & $0.2500$ & $0.9375$ & $\approx 0.2582$ \\
\bottomrule
\end{tabular}
\end{table}

Note that these constructed models are from scratch, with no relation to the
actual models we have trained beyond the square-wave structure and the
$\phi_{\out} = \phi_a + \phi_b$ relation. In \Cref{sec:extraction}, we
bridge this gap by using parameters extracted directly from the trained
models rather than constructing them ourselves.

\subsection{Extraction}
\label{sec:extraction}

To determine if there is latent structure in the weights of our noisy
models, we use a procedure to replace all $W_a\ti$, $W_b\ti$, and $W_\out\ti$
with square or cosine waves. The process is simple: For each neuron, we extract
the frequency from the dominant Fourier component of $W_a\ti$, and the phase
from the dominant Fourier component of $W_a\ti$, $W_b\ti$, $W_\out\ti$. Let
$\nu$ be the extracted frequency and $\phi_a\ti, \phi_b\ti, \phi_\out\ti$ be
the extracted phases. Then, we construct a model whose input and output weights
follow the extracted phases and frequencies. More formally,
\[
    W_a\ti[j] = f\left(\frac{2 \pi \nu j}{97} - \phi_a\ti\right),
    W_b\ti[j] = f\left(\frac{2 \pi \nu j}{97} - \phi_b\ti\right), W_\out\ti[j] = g\left(\frac{2 \pi \nu j}{97} - \phi_\out\ti\right). \\
\]
We test the same three settings for $f$ and $g$ as previously described. We
perform the extraction on actual models with various $\alpha$ values, using
ReLU activations, on both the final checkpoint and the saturation
checkpoint. We evaluate each extracted model on the validation set; our
results are summarized in
\Cref{tab:extraction_final,tab:extraction_saturation}.

\begin{table}[h]
\centering
\caption{Final checkpoint validation accuracy of real and extracted models across noise levels. Results are mean $\pm$ standard deviation over 15 seeds.}
\label{tab:extraction_final}
\begin{tabular}{c|cccc}
\toprule
$\alpha$ & Real & \coscos & \sqcos & \sqsq \\
\midrule
0.00 & $0.9998\pm0.0001$ & $\mathbf{0.9996\pm0.0006}$ & $0.9992\pm0.0005$ & $0.9840\pm0.0042$ \\
0.05 & $0.9270\pm0.0139$ & $\mathbf{0.9983\pm0.0009}$ & $0.9977\pm0.0010$ & $0.9723\pm0.0054$ \\
0.10 & $0.5518\pm0.0334$ & $0.9937\pm0.0045$ & $\mathbf{0.9945\pm0.0023}$ & $0.9586\pm0.0115$ \\
0.15 & $0.2070\pm0.0273$ & $0.9916\pm0.0033$ & $\mathbf{0.9920\pm0.0024}$ & $0.9522\pm0.0061$ \\
0.20 & $0.0523\pm0.0093$ & $0.9815\pm0.0063$ & $\mathbf{0.9847\pm0.0039}$ & $0.9326\pm0.0138$ \\
0.25 & $0.0120\pm0.0027$ & $0.9668\pm0.0097$ & $\mathbf{0.9742\pm0.0062}$ & $0.9108\pm0.0149$ \\
0.30 & $0.0040\pm0.0011$ & $0.9432\pm0.0174$ & $\mathbf{0.9582\pm0.0100}$ & $0.8875\pm0.0232$ \\
\bottomrule
\end{tabular}
\end{table}
\begin{table}[h]
\centering
\caption{Saturation checkpoint validation accuracy of real and extracted models across noise levels. Results are mean $\pm$ standard deviation over 15 seeds.}
\label{tab:extraction_saturation}
\begin{tabular}{c|cccc}
\toprule
$\alpha$ & Real & \coscos & \sqcos & \sqsq \\
\midrule
0.00 & $0.0000\pm0.0000$ & $0.8097\pm0.0407$ & $\mathbf{0.8520\pm0.0302}$ & $0.7359\pm0.0361$ \\
0.05 & $0.0000\pm0.0000$ & $0.8047\pm0.0479$ & $\mathbf{0.8490\pm0.0383}$ & $0.7361\pm0.0448$ \\
0.10 & $0.0001\pm0.0001$ & $0.7913\pm0.0544$ & $\mathbf{0.8404\pm0.0435}$ & $0.7287\pm0.0458$ \\
0.15 & $0.0002\pm0.0001$ & $0.7807\pm0.0464$ & $\mathbf{0.8287\pm0.0336}$ & $0.7099\pm0.0417$ \\
0.20 & $0.0004\pm0.0002$ & $0.7230\pm0.0828$ & $\mathbf{0.7803\pm0.0705}$ & $0.6644\pm0.0729$ \\
0.25 & $0.0004\pm0.0003$ & $0.6686\pm0.0639$ & $\mathbf{0.7285\pm0.0549}$ & $0.6069\pm0.0590$ \\
0.30 & $0.0010\pm0.0003$ & $0.6142\pm0.0679$ & $\mathbf{0.6760\pm0.0627}$ & $0.5618\pm0.0562$ \\
\bottomrule
\end{tabular}
\end{table}

The \sqcos extraction consistently achieves high validation accuracy on
final checkpoints across all noise levels, reaching 95.8\% on average at
$\alpha = 0.30$ where the real model achieves only 0.40\%. At lower noise
levels, \sqcos and \coscos perform similarly on the final checkpoint,
suggesting that for well-trained models the output phase structure is
sufficiently preserved that either input wave type recovers the algorithm.
The advantage of \sqcos becomes pronounced in the final checkpoints at high
noise ($\alpha \geq 0.25$) and is consistent across all saturation
checkpoints, where the real model accuracy is near zero throughout. This is
not a bias introduced by the extraction mechanism, as applying it to
completely random weights in place of the actual model weights results in
roughly $\frac{1}{97}$ accuracy, as expected from a random model. Thus the
saturation results confirm that the algorithmic structure---the appropriate
phases and frequencies to perform modular addition---is already encoded to a
significant degree at memorization, before grokking occurs.

We set all amplitudes to 1 and biases to 0 in the discussed extraction. To
determine the effect of learned amplitudes and biases on the learned
algorithm, we perform an ablation, performing the extraction using
amplitudes and biases extracted from the trained models. We refer to this as
the \textit{detailed extraction}. We find that the general trend remains
unchanged: \sqcos and \coscos perform similarly on final checkpoints at low
$\alpha$ levels, while \sqcos scores significantly higher on final
checkpoints with high $\alpha$, and across all saturation checkpoints. This
confirms that the amplitudes and biases do not have a bearing on the
qualitative results. We find that the accuracy in the detailed extractions
is comparable to that of the original extractions on final checkpoints, but
significantly lower on the saturation checkpoints (ranging from
$0.37$--$0.63$ as opposed to $0.67$--$0.86$). This suggests that the
amplitudes and biases during saturation are not yet fully suitable for
grokking, as not enough training has elapsed for the amplitudes to become
uniform and biases to become small in magnitude, as observed in the final
checkpoints. Numerical results for detailed extraction are provided in
Appendix~\ref{apx:extraction_amp}.

\section{Ablations}
\label{sec:ablations}

We use 3 independent random seeds (0, 1, and 2) per ablation. We fix $\alpha =
0.00$ in all ablations, so no label noise is applied.

\subsection{Activation function dependence}
\label{sec:activation_ablation}

To determine whether our findings are specific to ReLU or common across
activation functions, we repeat the same training procedure (identical
architecture, data, optimizer, and regularization) for GELU, Tanh, SiLU, and
Quadratic activations. For each activation we train three random seeds and
analyze both final and saturation checkpoints using the same periodicity
threshold ($>7.5$) on $W_a\ti$ to identify structured neurons.
\Cref{tab:activation_compact} summarizes key metrics for the final
checkpoint. All activations produce a substantial number of structured
neurons, and for ReLU, GELU, SiLU, and Quadratic the phase‑sum circular
correlation $r_{\mathrm{high}}$ among structured neurons exceeds $0.999$
(after dead neurons with near-zero weights are excluded, as described in
Appendix~\ref{apx:activation_full}; if these neurons are included in the
calculation, they introduce noise and the Quadratic activation model only
achieves $r_{\mathrm{high}} = 0.940$). This indicates near‑perfect adherence
to $\phi_\out = \phi_a + \phi_b$.

\begin{table}[h]
\scriptsize
\centering
\caption{Activation function ablation results.}
\label{tab:activation_compact}
\begin{tabular}{lccccc}
\toprule

Activation & Structured Neurons & $r_{\text{high}}$ & Val Acc & \sqcos (final) & \sqcos (saturation) \\
\midrule

relu & 233.3$\pm$3.8 & 0.999$\pm$0.000 & 1.000$\pm$0.000 & 0.999$\pm$0.000 & 0.849$\pm$0.032 \\
gelu & 228.3$\pm$4.9 & 1.000$\pm$0.000 & 0.999$\pm$0.000 & 0.997$\pm$0.001 & 0.884$\pm$0.013 \\
tanh & 251.3$\pm$2.5 & 0.048$\pm$0.074 & 0.996$\pm$0.002 & 0.008$\pm$0.001 & 0.010$\pm$0.001 \\
silu & 241.7$\pm$1.2 & 0.999$\pm$0.001 & 0.994$\pm$0.000 & 0.995$\pm$0.001 & 0.785$\pm$0.023 \\
quadratic & 151.3$\pm$5.0 & 0.940$\pm$0.089 & 0.996$\pm$0.002 & 1.000$\pm$0.000 & 1.000$\pm$0.000 \\
\bottomrule

\end{tabular}
\end{table}

\begin{figure}[t]
    \centering
    \includegraphics[width=\linewidth]{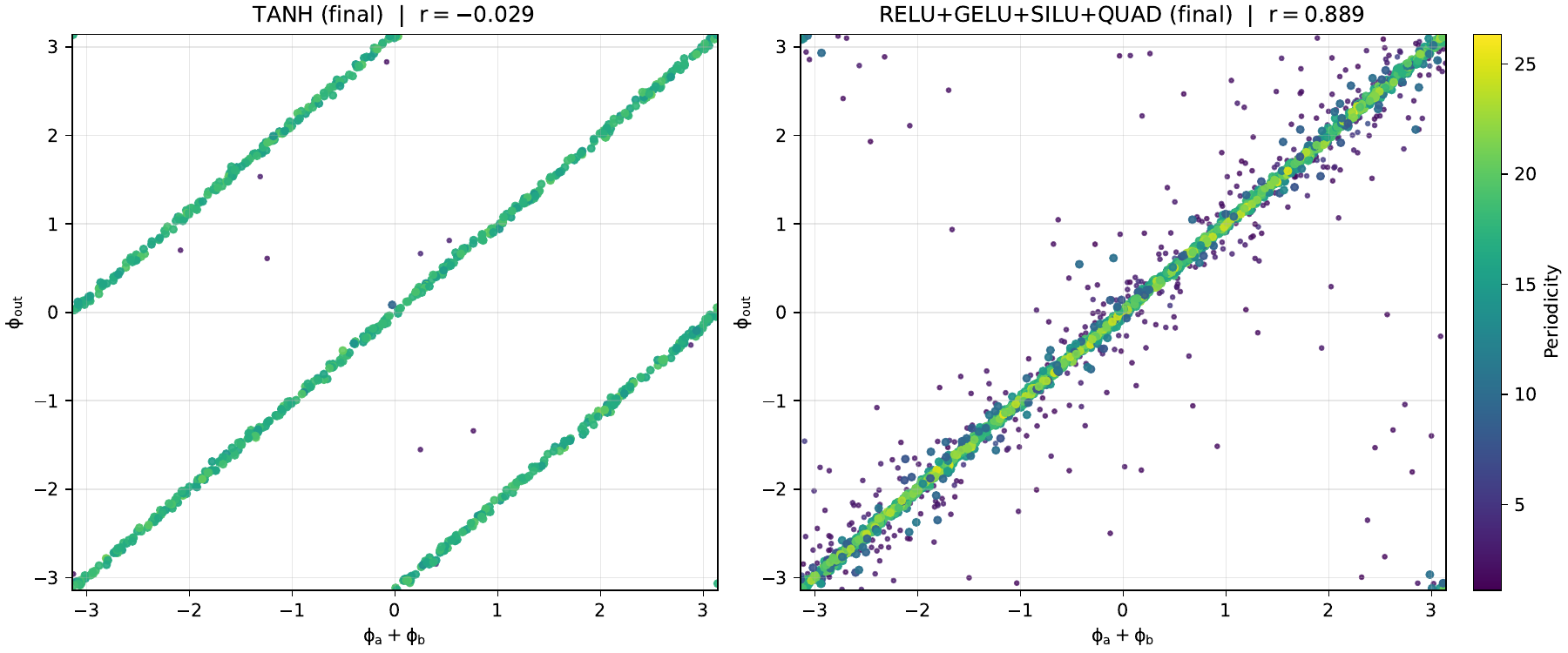}
    \caption{Final checkpoint $W_a\ti$ phase scatter for Tanh (left) vs.\ pooled ReLU, GELU, SiLU, and
    Quadratic (right), final checkpoints.}
    \label{fig:tanh_vs_pooled}
\end{figure}

Tanh is a notable exception: its $r_{\mathrm{high}}$ is near zero because
its phases appear to follow the relation $\phi_{\out} = \phi_a + \phi_b$ up
to shifts of multiples of $\pi$ rather than $2\pi$. The modulo‑$\pi$
circular correlation over all structured tanh neurons is $0.9674$, showing
that significant phase structure does exist. \Cref{fig:tanh_vs_pooled}
contrasts the phase scatter of Tanh with the neurons from the other four
activations. The two diagonals on the left plot represent $\phi_{\out} =
\phi_a + \phi_b$ and $\phi_{\out} = \phi_a + \phi_b + \pi$. Together, these
results demonstrate that the phase‑sum constraint, underlying algorithm, and
latent structure at saturation are robust across a range of activation
functions (ReLU, GeLU, SiLU, and Quadratic). Evidently, the algorithm
encoded in the Tanh model is somewhat different; we defer a more detailed
analysis of this to future work.

Quadratic activation exhibits a large fraction ($\sim 47\%$) of neurons with all
input weights zero to within $10^{-4}$; when these dead neurons are excluded, we
find no change in accuracy (see Appendix~\ref{apx:activation_full}). The \sqcos
extraction accuracy on the final checkpoint is high ($>0.96$) for all
activations except Tanh, and even at the saturation checkpoint \sqcos accuracy
remains well above chance for ReLU, GELU, SiLU, and Quadratic ($0.78$--$1.00$),
reinforcing the conclusion that the algorithmic structure is already present
before grokking.

We find that the neuron weights across all activations have a clear split
between high-periodicity and low-periodicity neurons, and that
high-periodicity neurons consistently tend to have higher output weight
norms and closely follow the phase-sum relation. Although the non-ReLU
activations do not result in square-wave weights, applying the \sqcos
extraction on all but Tanh retains near-perfect accuracy, and applying it to
saturation checkpoints recovers very high accuracy. This suggests that the
phase-sum relation is consistent across all activations as an optimal
strategy to solve modular addition, while the square-wave pattern is
specific to ReLU activations despite being capable of high accuracy on other
activations. We provide a more detailed discussion of these results in
Appendix~\ref{apx:activation_full}.

\subsection{Model Parameters}
\label{sec:param}

We ablate over the hidden neuron count $N$ ($N \in \{16, 32, 64, 96, 128, 192,
256, 384, 512\}$), keeping $p$ fixed at $97$. For model widths that successfully
reach high accuracy, namely $N \ge 96$, we find that square waves emerge on all
seeds, a bimodal periodicity distribution exists, and high-periodicity neurons
follow the phase-sum relation almost exactly. We find that as $N$ increases,
bimodality of input weights increases, and the $3k$ and $5k$ harmonic Fourier
weight magnitude ratios relative to the fundamental frequency weight magnitude
increase (to $0.22$ and $0.07$ respectively) while all others remain quite low
($0.04$--$0.06$). This suggests that with higher $N$, the weights are more
closely aligned with the theoretical square waves.

We also ablate over modulus $p$ (we test $p \in \{61, 73, 80, 96, 101, 102,
149\}$), with $N$ constant at $256$. All tested $p$ values except $61$
achieve near-perfect at this width, and we observe square waves in the input
weights: The bimodality across all tested $p$ is nearly identical to that of
the $p = 97$ model, and the $3k$ and $5k$ harmonic weight magnitudes are
consistently larger than other harmonic magnitudes. The bimodal periodicity
distribution and high adherence to $\phi_{\out} = \phi_a + \phi_b$ holds as
well.

Across all $N \ge 96$ and all tested $p$, the extracted models from the
final checkpoint achieve near-$100\%$ accuracy. These ablations show that
the square-wave finding and the extraction generalize to larger hidden
widths and other moduli. Full results and a more detailed discussion are
provided in Appendix~\ref{apx:param}.

\section{Discussion}

Our findings suggest a reframing of what grokking accomplishes. Prior
mechanistic studies have shown that grokking involves a gradual circuit
formation phase during memorization, where structured mechanisms amplify
continuously before the test accuracy jump \citep{nanda2023}. Our results extend
this picture in ReLU MLPs under label noise: We extract a high-accuracy
idealized model directly from memorization-phase (saturation) weights, even in
non-grokking regimes where the real model fails to generalize. This demonstrates
that the core algorithmic structure---frequencies and phase-sum relations---is
encoded in a form sufficient for near-recovery of the solution as early as
memorization, and persists robustly under corruption. What grokking does is
sharpen this latent structure, aligning model weights more closely with the
ideal waves.

Our finding that input weights follow square waves rather than sinusoidal waves
appears to be unique to ReLU and not covered in any previous work.
\citet{gromov2023} analytically construct cosine weights and find that periodic
structure appears in ReLU MLPs trained on modular addition with vanilla gradient
descent and no regularization, but do not demonstrate whether the weights follow
sinusoidal or square waves; \citet{doshi2024} use AdamW with weight decay but
primarily study quadratic activations. Our setting uses ReLU activations with
AdamW and weight decay. It appears from our ablations that across different
activation functions, the phase-sum relation $\phi_{\out} = \phi_a + \phi_b
\bmod 2\pi$ (in the case of ReLU, SiLU, GeLU, and Quadratic activations) or mod
$\pi$ (in the case of Tanh) arises when grokking, while the square-wave
phenomenon appears only in the case of ReLU. We demonstrate through ablations
that these results are robust to different moduli and hidden widths.

\textbf{Limitations.} Our experiments are confined to a single task (modular
addition) and a single architecture (one-hidden-layer MLP). Whether the latent structure finding
generalizes to other tasks or architectures is an important direction for
future work. Additionally, we
do not analyze the precise structure of the output layer weights beyond
periodicity and frequency matching.

\section{Conclusion}

We have shown that ReLU MLPs trained on modular addition learn near-binary
square wave input weights whose dominant Fourier phases satisfy $\phi_{\out} =
\phi_a + \phi_b$, and that this phase structure is largely preserved even under
heavy label noise that prevents generalization. Extracting these phases and
constructing idealized models recovers 95.8\% accuracy from models that
themselves achieve 0.40\%, and substantial accuracy from saturation checkpoints
where the real model has not yet generalized at all. Together, these results
suggest that grokking does not discover the correct algorithm---it sharpens one
that was already there.

\newpage

\bibliographystyle{plainnat}
\bibliography{paper.bib}

\newpage

\appendix

\section{Theoretical Proofs}
\label{apx:theory}

In \Cref{sec:theory}, we introduced idealized multi-layer perceptrons
constructed from square and cosine waves to investigate the algorithmic
structure learned by the network. Here, we formally prove that all three
idealized constructions (\coscos, \sqcos, \sqsq) almost surely achieve
perfect accuracy on the modular addition task as the hidden layer width $N$
approaches infinity. Furthermore, we derive the exact expected margin $M$
and variance $V$ values presented in Table 2, which provide signal-to-noise
ratios that act as a theoretical measure of performance of the constructed
models.

\subsection{Framework and Definitions}
\label{apx:theory:framework}

\textbf{Setup.} Recall that the network is trained on the modular addition task
$a + b \equiv c^* \pmod p$, where $p$ is the modulus, and the integer inputs $a,
b \in \{0, 1, \dots, p-1\}$ are provided via concatenated one-hot encodings. Let
$N$ denote the width of the hidden layer (number of hidden neurons) with a ReLU
activation function.

For each neuron $i \in \{0, 1, \dots, N - 1\}$, we define independent phase offsets
$\phi_a\ti, \phi_b\ti \sim \mathcal{U}[-\pi, \pi]$. We define the variables
\[
    A\ti = \frac{2\pi ia}{p} - \phi_a\ti, \qquad B\ti = \frac{2\pi ib}{p} - \phi_b\ti, \qquad C\ti_c = \frac{2 \pi ic}{p} - \phi_{\out}\ti.
\]
We consider $A\ti$, $B\ti$, and $C\ti_c$ modulo $2\pi$. Because $\phi_a\ti$ and
$\phi_b\ti$ are uniformly distributed, $A\ti, B\ti$ must be uniform in the range
$[-\pi, \pi]$. Note that $f(A\ti) + f(B\ti)$ is the preactivation for neuron $i$
and $g(C\ti_c)$ is the weight between hidden neuron $i$ and output class $c$.

Note that
\begin{align*}
    C_c\ti &= \frac{2 \pi ic}{p} - \phi_{\out}\ti \\
           &= \frac{2 \pi ic}{p} - \phi_a\ti - \phi_b\ti \\
           &= \frac{2 \pi ia}{p} + \frac{2 \pi ib}{p} + \frac{2 \pi i(c - a - b)}{p} - \phi_a\ti - \phi_b\ti \\
           &= A\ti + B\ti + \frac{2 \pi (c - a - b)i}{p}.
\end{align*}
Let $\Delta\ti_c = \frac{2 \pi (c - a - b)i}{p}$ be this last term, so
$C_c\ti = A\ti + B\ti + \Delta\ti_c$.

As in \Cref{sec:ideal}, let $f$ denote the wave function for the input weights
(either $\cos$ or $\sign\circ\cos$) and $g$ denote the wave function for the
output weights. The logit $L_c$ for a given class $c \in \{0, 1, \dots, p-1\}$
is computed as the sum over all $N$ hidden neurons,
\[
    L_c = \sum_{i=0}^{N-1} X_c\ti \text{, where } X_c\ti = \relu\!\left(f(A\ti) + f(B\ti)\right) \cdot g\left(C_c\ti\right).
\]

Since the random variables $X_c\ti$ are independent across $i$ and bounded
(within the range $[-2, 2]$), the strong law of large numbers dictates that
\begin{equation}
    L_c = \sum_{i=0}^{N-1} \E\left[X_c\ti\right] + o(N) \quad\text{almost surely.}
    \label{eq:slln}
\end{equation}
For the correct class $c = c^* \equiv a + b \bmod p$, it is guaranteed that
$p \mid i(c^*-a-b)$, and hence $\Delta_{c^*}\ti \equiv 0 \pmod{2\pi}$ for
all $i$.

\textbf{Cyclic-sum bound.} For incorrect classes ($c \neq c^*$), we define
the offset $s \equiv c - a - b \equiv c - c^* \bmod p$. Because $1 \le s < p$, it is guaranteed
that $p \nmid s$. We will frequently rely on the following exponential sum
estimate to bound the incorrect logits:
\begin{equation}
    \left|\sum_{i=0}^{N-1} \cos\left(\frac{2\pi is}{p}\right)\right| \leq p - 1.
    \label{eq:blocksum}
\end{equation}
This holds because, if $N = mp + r$ for integer $m$ and integer $r \in [0, p
- 1]$, then
\begin{equation*}
    \sum_{i=0}^{N-1} \cos\left(\frac{2\pi is}{p}\right) = \sum_{i=0}^{r-1} \cos\left(\frac{2\pi is}{p}\right),
\end{equation*}
as $\sum_{i=0}^{p-1} \cos\left(\frac{2 \pi is}{p}\right) = 0$. Of course,
$\left|\cos\left(\frac{2\pi is}{p}\right)\right| \le 1$, so the absolute
value of the full sum is no more than $r \le p - 1$.

\subsection{\coscos}

We calculate the expected value of the logit contribution from a given neuron
$i$ to class $c$ (i.e.\ $\E[X_c\ti]$). For brevity, we omit the superscript
$\ti$, so, for example, in the following calculations, $A$ represents $A\ti$.

\begin{align*}
    &\frac{1}{4\pi^2} \int_{-\pi}^{\pi} \int_{-\pi}^{\pi} \relu(\cos(A) + \cos(B)) \cos(C_c)\,dA\,dB \\
    = &\frac{1}{4\pi^2} \int_{-\pi}^{\pi} \int_{-\pi}^{\pi} \relu(\cos(A) + \cos(B)) \cos(A + B + \Delta_c)\,dA\,dB \\
    = &\frac{1}{4\pi^2} \int_{-\pi}^{\pi} \int_{-\pi}^{\pi} \relu\left(2\cos\left(\frac{A+B}{2}\right)\cos\left(\frac{A-B}{2}\right)\right) \cos(A + B + \Delta_c)\,dA\,dB \\
    = &\frac{1}{2\pi^2} \int_{-\pi}^{\pi} \int_{-\pi}^{\pi} \relu\left(\cos\left(\frac{A+B}{2}\right)\cos\left(\frac{A-B}{2}\right)\right) \cos(A + B + \Delta_c)\,dA\,dB,
\end{align*}
by the sum-to-product identity. Note that we need only consider the case where
$\cos\left(\frac{A+B}{2}\right)\cos\left(\frac{A-B}{2}\right)$ is positive, as
ReLU zeros out the integrand when this value is negative.

By inspection, there are no values of $A$ and $B$ such that both factors are
negative, so this value is positive if and only if both cosine factors are
positive, i.e.\ when $\frac{A+B}{2}$ and $\frac{A-B}{2}$ both lie in
$(-\frac{\pi}{2}, \frac{\pi}{2})$. Thus, substituting $u = \frac{A+B}{2}$ and $v
= \frac{A-B}{2}$, we have
\begin{align*}
    &\frac{1}{2\pi^2} \int_{-\pi}^{\pi} \int_{-\pi}^{\pi} \relu\left(\cos\left(\frac{A+B}{2}\right)\cos\left(\frac{A-B}{2}\right)\right) \cos(A + B + \Delta_c)\,dA\,dB \\
    = &\frac{1}{2\pi^2} \int_{-\frac{\pi}{2}}^{\frac{\pi}{2}} \int_{-\frac{\pi}{2}}^{\frac{\pi}{2}} \relu\left(\cos(u)\cos(v)\right) \cos(2u + \Delta_c)(2\,du\,dv) \\
    = &\frac{1}{\pi^2} \int_{-\frac{\pi}{2}}^{\frac{\pi}{2}} \int_{-\frac{\pi}{2}}^{\frac{\pi}{2}} \cos(u)\cos(v) \cos(2u + \Delta_c)\,du\,dv \\
    = &\frac{1}{\pi^2} \int_{-\frac{\pi}{2}}^{\frac{\pi}{2}} \cos(v) \int_{-\frac{\pi}{2}}^{\frac{\pi}{2}} \cos(u) \cos(2u + \Delta_c)\,du\,dv \\
    = &\frac{1}{\pi^2} \left(\int_{-\frac{\pi}{2}}^{\frac{\pi}{2}} \cos(v)\,dv\right) \left(\int_{-\frac{\pi}{2}}^{\frac{\pi}{2}} \frac12 \left(\cos(3u + \Delta_c) + \cos(u + \Delta_c)\right)\,du\right), \\
\end{align*}
by the product-to-sum identity. Evaluating this final expression yields
$\E[X_c\ti] = \frac{4}{3\pi^2}\cos(\Delta_c\ti)$.

\textbf{Correctness.} For $c = c^*$, $\Delta_c\ti = 0$, so $\E[X_{c^*}\ti]
= \frac{4}{3\pi^2}$, giving $L_{c^*} = \frac{4}{3\pi^2}N + o(N)$ almost surely.
For $c \neq c^*$, $s = c - a - b$ satisfies $p \nmid s$, so by~\eqref{eq:slln}
and~\eqref{eq:blocksum},
\[
    L_c = \frac{4}{3\pi^2}\sum_{i=0}^{N-1}\cos\!\left(\frac{2\pi is}{p}\right) + o(N), \qquad\left|L_c\right| \leq \frac{4}{3\pi^2}(p-1) + o(N).
\]
Hence, $L_{c^*} - L_c \ge \frac{4}{3\pi^2}(N - p + 1) + o(N)$ which approaches
$\infty$ almost surely as $N$ approaches $\infty$. We conclude that $\argmax_c
L_c = c^*$ for sufficiently large $N$ almost surely.

\textbf{Variance.} We now calculate the variance $V_c\ti = \E[(X_c\ti)^2]
- \E[X_c\ti]^2$. Using the same simplifications and substitutions as before,
we arrive at
\begin{align*}
    \E[(X_c\ti)^2] = &\frac{1}{4\pi^2} \int_{-\pi}^{\pi} \int_{-\pi}^{\pi} \left(\relu(\cos(A) + \cos(B)) \cos(C_c)\right)^2\,dA\,dB \\
    = &\frac{1}{4\pi^2} \int_{-\frac{\pi}{2}}^{\frac{\pi}{2}} \int_{-\frac{\pi}{2}}^{\frac{\pi}{2}} \left(2\cos(u)\cos(v) \cos(2u + \Delta_c)\right)^2(2\,du\,dv) \\
    = &\frac{2}{\pi^2} \left(\int_{-\frac{\pi}{2}}^{\frac{\pi}{2}} \cos^2(v)\,dv\right) \left(\int_{-\frac{\pi}{2}}^{\frac{\pi}{2}} \left(\cos(u) \cos(2u + \Delta_c)\right)^2\,du\right) \\
    = &\frac{2}{\pi^2} \left(\int_{-\frac{\pi}{2}}^{\frac{\pi}{2}} \cos^2(v)\,dv\right) \left(\int_{-\frac{\pi}{2}}^{\frac{\pi}{2}} \left(\frac12 (\cos(3u + \Delta_c) + \cos(u + \Delta_c))\right)^2\,du\right) \\
    = &\frac{1}{2\pi^2} \left(\int_{-\frac{\pi}{2}}^{\frac{\pi}{2}} \cos^2(v)\,dv\right) \left(\int_{-\frac{\pi}{2}}^{\frac{\pi}{2}} \left(\cos^2(3u + \Delta_c) + \cos^2(u + \Delta_c) + 2\cos(3u + \Delta_c)\cos(u + \Delta_c)\right)\,du\right) \\
    = &\frac{1}{2\pi^2} \left(\int_{-\frac{\pi}{2}}^{\frac{\pi}{2}} \cos^2(v)\,dv\right) \left(\int_{-\frac{\pi}{2}}^{\frac{\pi}{2}} \left(\cos^2(3u + \Delta_c) + \cos^2(u + \Delta_c) + \cos(4u + 2\Delta_c) + \cos(2u)\right)\,du\right) \\
    = &\frac{1}{2\pi^2} \left(\frac{\pi}{2}\right) \left(\frac{\pi}{2} + \frac{\pi}{2} + 0 + 0\right) = \frac{1}{4}.
\end{align*}
Indeed, the expectation of the square evaluates to a constant regardless of
the value of $\Delta_c\ti$. Thus we find that the variance $V_c\ti$ is
$\frac{1}{4} - \frac{16}{9\pi^4}\cos^2(\Delta_c\ti)$, and this is
$\frac{1}{4} - \frac{16}{9\pi^4} \approx 0.232$ for $c = c^*$ (where
$\Delta_c\ti = 0$).

\subsection{\sqcos}

\textbf{Expected value.} In this case,
\[
    X_c\ti = \relu(\sign(\cos(A\ti)) + \sign(\cos(B\ti))) \cos(A\ti + B\ti + \Delta_c\ti).
\]
We will compute
\[
    \E[X_c\ti] = \frac{1}{4\pi^2} \int_{-\pi}^{\pi} \int_{-\pi}^{\pi} \relu(\sign(\cos(A)) + \sign(\cos(B))) \cos(A + B + \Delta)\,dA\,dB.
\]

Since there are only finitely many values for $A$ that make $\cos(A)$ (and
consequently $\sign(\cos(A))$) equal to 0, we may ignore these cases and
consider only $\sign(\cos(A)) = \pm 1$, and likewise for $\sign(\cos(B))$. The
expression within the $\relu$ can evaluate to either $2$, $0$, or $-2$. The
latter two cases make the integrand zero after ReLU, so we need only consider
the case when $\sign(\cos(A)) + \sign(\cos(B)) = 2$. This is true if and only if
$A, B \in (-\frac{\pi}{2}, \frac{\pi}{2})$. Thus, we have
\begin{align*}
    &\frac{1}{4\pi^2} \int_{-\pi}^{\pi} \int_{-\pi}^{\pi} \relu(\sign(\cos(A)) + \sign(\cos(B))) \cos(A + B + \Delta)\,dA\,dB \\
    = &\frac{1}{4\pi^2} \int_{-\frac{\pi}{2}}^{\frac{\pi}{2}} \int_{-\frac{\pi}{2}}^{\frac{\pi}{2}} \relu(2) \cos(A + B + \Delta)\,dA\,dB \\
    = &\frac{1}{2\pi^2} \int_{-\frac{\pi}{2}}^{\frac{\pi}{2}} \int_{-\frac{\pi}{2}}^{\frac{\pi}{2}} \cos(A + B + \Delta)\,dA\,dB \\
    = &\frac{1}{2\pi^2} \cdot 4\cos(\Delta) = \frac{2}{\pi^2} \cos(\Delta_c\ti).
\end{align*}

\textbf{Correctness.} Similarly to the previous case, for $c = c^*$, $L_{c^*} =
\frac{2}{\pi^2}N + o(N)$ almost surely, and for $c \ne c^*$, $|L_c| \le
\frac{2}{\pi^2}(p - 1) + o(N)$ almost surely. Thus $L_{c^*} - L_c \ge
\frac{2}{\pi^2} (N - p + 1) + o(N)$ almost surely, which approaches $\infty$ as
$N$ approaches $\infty$. Therefore, with sufficiently large $N$, $\argmax_c L_c
= c^*$ almost surely.

\textbf{Variance.} We now calculate the variance $V_c\ti = \E[(X_c\ti)^2] - \E[X_c\ti]^2$.
\begin{align*}
    \E[(X_c\ti)^2] &= \frac{1}{4\pi^2} \int_{-\pi}^{\pi} \int_{-\pi}^{\pi} (\relu(\sign(\cos(A)) + \sign(\cos(B))) \cos(A + B + \Delta))^2\,dA\,dB \\
                   &= \frac{1}{4\pi^2} \int_{-\frac{\pi}{2}}^{\frac{\pi}{2}} \int_{-\frac{\pi}{2}}^{\frac{\pi}{2}} (2 \cos(A + B + \Delta))^2\,dA\,dB \\
                   &= \frac{1}{\pi^2} \int_{-\frac{\pi}{2}}^{\frac{\pi}{2}} \int_{-\frac{\pi}{2}}^{\frac{\pi}{2}} \cos^2(A + B + \Delta)\,dA\,dB \\
                   &= \frac{1}{\pi^2} \cdot \frac{\pi^2}{2} = \frac{1}{2}.
\end{align*}
Once again, this evaluates to a constant regardless of the value of
$\Delta_c\ti$. Thus, $V_c\ti = \frac{1}{2} -
\frac{4}{\pi^4}\cos^2(\Delta_c\ti)$, which evaluates to $\frac{1}{2} -
\frac{4}{\pi^4} \approx 0.459$ for the correct class $c^*$.

\subsection{\sqsq}

\textbf{Expected value.} In this case,
\[
    X_c\ti = \relu(\sign(\cos(A\ti)) + \sign(\cos(B\ti))) \sign(\cos(A\ti + B\ti + \Delta_c\ti)).
\]
We will compute
\[
    \E[X_c\ti] = \frac{1}{4\pi^2} \int_{-\pi}^{\pi} \int_{-\pi}^{\pi} \relu(\sign(\cos(A)) + \sign(\cos(B))) \sign(\cos(A + B + \Delta))\,dA\,dB.
\]

By the same logic as in the previous case, we need consider only the case where
$A, B \in (-\frac{\pi}{2}, \frac{\pi}{2})$.
\begin{align*}
    &\frac{1}{4\pi^2} \int_{-\pi}^{\pi} \int_{-\pi}^{\pi} \relu(\sign(\cos(A)) + \sign(\cos(B))) \sign(\cos(A + B + \Delta))\,dA\,dB \\
    = &\frac{1}{4\pi^2} \int_{-\frac{\pi}{2}}^{\frac{\pi}{2}} \int_{-\frac{\pi}{2}}^{\frac{\pi}{2}} 2 \sign(\cos(A + B + \Delta))\,dA\,dB \\
    = &\frac{1}{2\pi^2} \int_{-\frac{\pi}{2}}^{\frac{\pi}{2}} \int_{-\frac{\pi}{2}}^{\frac{\pi}{2}} \sign(\cos(A + B + \Delta))\,dA\,dB.
\end{align*}

We perform the substitution $u = A + B$, $v = A$. The Jacobian, $\begin{bmatrix}
    1 & 1 \\
    1 & 0
\end{bmatrix}$, has determinant $-1$, so we have
\begin{align*}
    &\frac{1}{2\pi^2} \int_{-\frac{\pi}{2}}^{\frac{\pi}{2}} \int_{-\frac{\pi}{2}}^{\frac{\pi}{2}} \sign(\cos(A + B + \Delta))\,dA\,dB \\
    = &\frac{1}{2\pi^2} \int_{-\pi}^{\pi} \int_{\max(-\frac{\pi}{2},\, u - \frac{\pi}{2})}^{\min(\frac{\pi}{2},\, u + \frac{\pi}{2})} \sign(\cos(u + \Delta))\,dv\,du \\
    = &\frac{1}{2\pi^2} \int_{-\pi}^{\pi} (\pi - |u|) \sign(\cos(u + \Delta))\,du,
\end{align*}
because $\min(\frac{\pi}{2}, u + \frac{\pi}{2}) - \max(-\frac{\pi}{2}, u -
\frac{\pi}{2}) = \pi - |u|$. Since $\int_{-\pi}^{\pi} \sign(\cos(u +
\Delta))\,du = 0$ for any $\Delta$ (the positive and negative half-periods of
cosine have equal measure), we may replace
$\pi - |u|$ by $-|u|$:
\[
    \E[X_c\ti] = -\frac{1}{2\pi^2} \int_{-\pi}^{\pi} |u|\,\sign(\cos(u + \Delta_c\ti))\,du.
    \label{eq:sqsq-exp}
\]
For $|\Delta| \leq \frac{\pi}{2}$, the set $\{u \in [-\pi, \pi] : \cos(u +
\Delta) > 0\} = (-\frac{\pi}{2} - \Delta,\, \frac{\pi}{2} - \Delta)$ lies
entirely within $(-\pi, \pi)$. Writing the integral as twice the integral over
the positive region minus the integral over all of $[-\pi, \pi]$,
\begin{align*}
    \int_{-\pi}^{\pi} |u|\,\sign(\cos(u + \Delta))\,du
    &= 2\int_{-\frac{\pi}{2}-\Delta}^{\frac{\pi}{2}-\Delta} |u|\,du - \int_{-\pi}^{\pi} |u|\,du \\
    &= \left(\left(\tfrac{\pi}{2}+\Delta\right)^2 + \left(\tfrac{\pi}{2}-\Delta\right)^2\right) - \pi^2 \\
    &= 2\Delta^2 - \frac{\pi^2}{2} \\
    \implies \E[X_c\ti] &= -\frac{1}{2\pi^2} \left(2\left(\Delta_c\ti\right)^2 - \frac{\pi^2}{2}\right),
\end{align*}
so $\E[X_c\ti] = \frac{1}{4} - \frac{(\Delta_c\ti)^2}{\pi^2}$ for $|\Delta_c\ti|
\leq \frac{\pi}{2}$. For $\Delta_c\ti \in [\frac{\pi}{2}, \frac{3\pi}{2}]$, by
evaluating as the integral over the full domain minus twice the integral over
the negative region, we find $\E[X_c\ti] = \frac{(\Delta_c\ti - \pi)^2}{\pi^2} -
\frac{1}{4}$. We let $f$ be a function such that $\E[X_c\ti] =
f(\Delta_c\ti)$. This is, of course, periodic, with period $2\pi$.
Additionally, it shares some properties with $\cos(x)$: $f(x + \pi) = -f(x)$
and $f(-x) = f(x)$.

\begin{lemma}
\label{lma:fsum}
For all positive integers $p$,
\[
    \sum_{n=0}^{p-1} f\!\left(\frac{2\pi n}{p}\right) = \begin{cases} 0 & \text{if $p$ even,} \\ \dfrac{(-1)^{(p-1)/2}}{4p^2} & \text{if $p$ odd.} \end{cases}
\]
\end{lemma}
\begin{proof}
Even $p$: Recall that $f$ satisfies $f(x) + f(x + \pi) = 0$. Pairing $n$ with $n + p/2$ yields $0$, so the sum is $0$.

Odd $p$: Using the symmetry $f(x) = f(2\pi - x)$ and $f(0) = \frac{1}{4}$,
\[
    \sum_{n=0}^{p-1} f\!\left(\frac{2\pi n}{p}\right) = \frac{1}{4} + 2\sum_{n=1}^{(p-1)/2} f\!\left(\frac{2\pi n}{p}\right).
\]
\begin{itemize}
    \item \textit{Case $p = 4k+1$.} For $n \in \{1, \dots, k\}$, $f\!\left(\frac{2\pi n}{p}\right) = \frac{1}{4} - \frac{4n^2}{p^2}$, and for $n \in \{k+1, \dots, 2k\}$, $f\!\left(\frac{2\pi n}{p}\right) = \frac{(2n-p)^2}{p^2} - \frac{1}{4}$. Applying $\sum_{n=1}^{k} n^2 = \frac{k(k+1)(2k+1)}{6}$ and $\sum_{j=1}^{k}(2j-1)^2 = \frac{k(2k-1)(2k+1)}{3}$,
    \begin{align*}
        \sum_{n=0}^{p-1} f\!\left(\frac{2\pi n}{p}\right)
        &= \frac{1}{4} + 2\left(\frac{k}{4} - \frac{2k(k+1)(2k+1)}{3p^2} + \frac{k(2k-1)(2k+1)}{3p^2} - \frac{k}{4}\right) \\
        &= \frac{1}{4} - \frac{2k(2k+1)}{p^2}.
    \end{align*}
    Substituting $k = \frac{p-1}{4}$ and $2k+1 = \frac{p+1}{2}$ gives $\frac{1}{4} - \frac{p^2-1}{4p^2} = \frac{1}{4p^2}$.
    \item \textit{Case $p = 4k+3$.} For $n \in \{1, \dots, k\}$, $f\!\left(\frac{2\pi n}{p}\right) = \frac{1}{4} - \frac{4n^2}{p^2}$, and for $n \in \{k+1, \dots, 2k+1\}$, $f\!\left(\frac{2\pi n}{p}\right) = \frac{(2n-p)^2}{p^2} - \frac{1}{4}$. Applying $\sum_{n=1}^{k} n^2 = \frac{k(k+1)(2k+1)}{6}$ and $\sum_{j=1}^{k+1}(2j-1)^2 = \frac{(k+1)(2k+1)(2k+3)}{3}$,
    \begin{align*}
        \sum_{n=0}^{p-1} f\!\left(\frac{2\pi n}{p}\right)
        &= \frac{1}{4} + 2\left(\frac{k}{4} - \frac{2k(k+1)(2k+1)}{3p^2} + \frac{(k+1)(2k+1)(2k+3)}{3p^2} - \frac{k+1}{4}\right) \\
        &= -\frac{1}{4} + \frac{2(k+1)(2k+1)}{p^2}.
    \end{align*}
    Substituting $k+1 = \frac{p+1}{4}$ and $2k+1 = \frac{p-1}{2}$ gives $-\frac{1}{4} + \frac{p^2-1}{4p^2} = -\frac{1}{4p^2}$.
\end{itemize}
Combining both subcases gives $\frac{(-1)^{(p-1)/2}}{4p^2}$.
\end{proof}

\begin{lemma}
\label{lma:fsumtotal}
For $s \in \{1, \dots, p-1\}$,
\[
    \left|\sum_{i=0}^{N-1} f\!\left(\frac{2\pi si}{p}\right)\right| \leq \frac{N}{32} + \frac{p-1}{4}.
\]
\end{lemma}
\begin{proof}
    Let $q = p/\gcd(s,p) \geq 2$ be the period of the sequence $i \mapsto \frac{2\pi si}{p} \pmod{2\pi}$. Writing $N = mq + r$ with $r \in [0, q-1]$ and applying \Cref{lma:fsum},
\[
    \sum_{i=0}^{N-1} f\!\left(\frac{2\pi si}{p}\right) = m\sum_{n=0}^{q-1} f\!\left(\frac{2\pi n}{q}\right) + \sum_{i=0}^{r-1} f\!\left(\frac{2\pi si}{p}\right) \le \frac{m}{4q^2} + \sum_{i=0}^{r-1} f\!\left(\frac{2\pi si}{p}\right).
\]
Since $|f| \leq \frac{1}{4}$ everywhere and $m \leq N/q$, the absolute value is at most $\frac{N}{4q^3} + \frac{q-1}{4} \leq \frac{N}{32} + \frac{p-1}{4}$.
\end{proof}

\textbf{Correctness.} For $c = c^*$, $\Delta_{c^*}\ti = 0$ for all $i$, so $\E[X_{c^*}\ti] = f(0) = \frac{1}{4}$, giving $L_{c^*} = \frac{N}{4} + o(N)$ almost surely. For $c \neq c^*$, $\E[X_c\ti] = f\!\left(\frac{2\pi si}{p}\right)$, so by Lemma~2 and~\eqref{eq:slln}, $|L_c| \leq \frac{N}{32} + \frac{p-1}{4} + o(N)$ almost surely. Therefore,
\[
    L_{c^*} - L_c \geq \frac{7N}{32} - \frac{p-1}{4} + o(N),
\]
which approaches $\infty$ almost surely as $N$ approaches $\infty$. We conclude that $\argmax_c L_c = c^*$ for sufficiently large $N$ almost surely.

\textbf{Variance.} Since $(\sign(\cos(x)))^2 = 1$ for almost all real $x$,
\[
    \E[(X_{c^*}\ti)^2] = \frac{1}{4\pi^2} \int_{-\frac{\pi}{2}}^{\frac{\pi}{2}} \int_{-\frac{\pi}{2}}^{\frac{\pi}{2}} (2 \sign(\cos(A + B + \Delta)))^2\,dA\,dB = \frac{1}{4\pi^2} \int_{-\frac{\pi}{2}}^{\frac{\pi}{2}} \int_{-\frac{\pi}{2}}^{\frac{\pi}{2}} 4\,dA\,dB = 1.
\]
Thus $V = \E[(X_{c^*}\ti)^2] - \E[X_{c^*}\ti]^2 = 1 - \frac{1}{16} =
\frac{15}{16} \approx 0.938$.

\newpage

\section{Periodicity Gap}
\label{apx:periodicity}

\begin{figure}[h]
    \centering
    \includegraphics[width=0.4\linewidth]{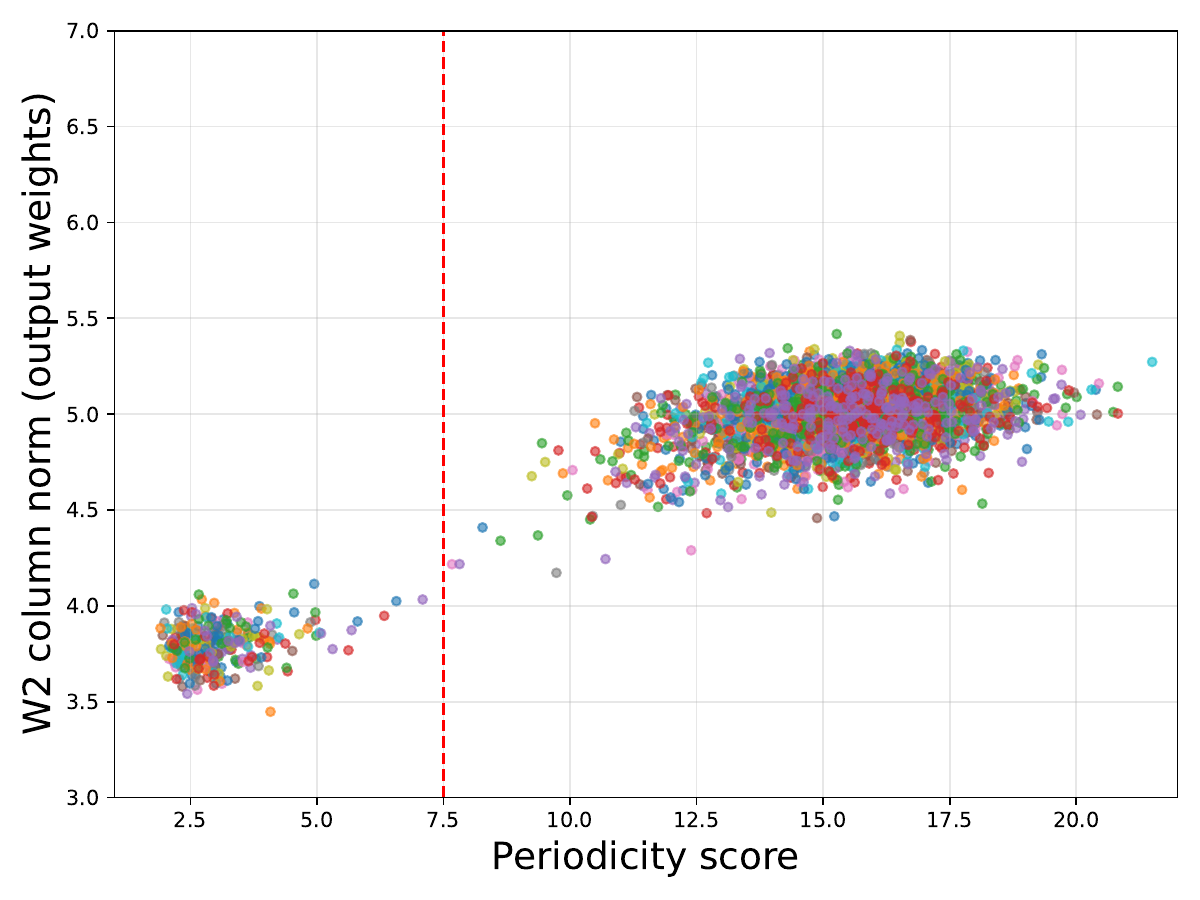}
    \caption{Periodicity vs.\ $L^2$ output weight norm over all
    neurons for $\alpha = 0.00$. Pooled over all 15 seeds. Each color
    represents a different seed. The red dashed line represents the
    periodicity threshold, $7.5$.}
    \label{fig:periodicity_alpha0.00}
\end{figure}

In the clean model, the periodicity of neurons follows a bimodal
distribution. In addition, the output weights for neurons in the
lower-periodicity group consistently have lower $L^2$ norms; this is clearly
visible in \Cref{fig:periodicity_alpha0.00}. The positive correlation
between periodicity and output weight norm shows that the low-periodicity
neurons have relatively little effect on the final answer, consistent with
the negligible drop in accuracy when these neurons are ablated.

We choose the threshold $7.5$ to separate the higher-periodicity group from
the lower-periodicity one, but this choice is not sensitive: any choice of
threshold in the range $[6, 9]$ yields nearly identical groupings, with no
more than 2 neurons per seed changing assignment as the threshold varies.

\begin{figure}[h]
    \centering
    \begin{subfigure}[b]{0.32\linewidth}
        \includegraphics[width=\linewidth]{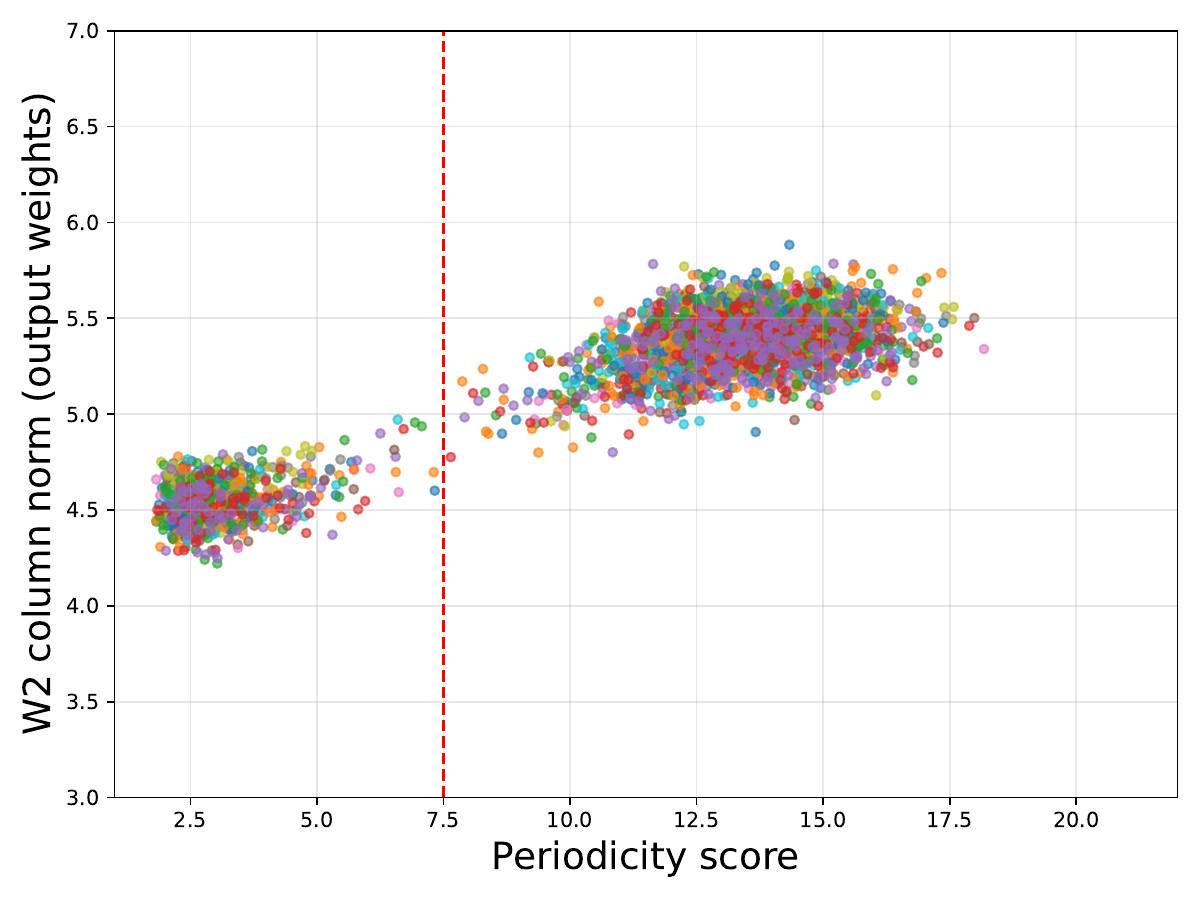}
        \caption{$\alpha = 0.05$}
    \end{subfigure}\hfill
    \begin{subfigure}[b]{0.32\linewidth}
        \includegraphics[width=\linewidth]{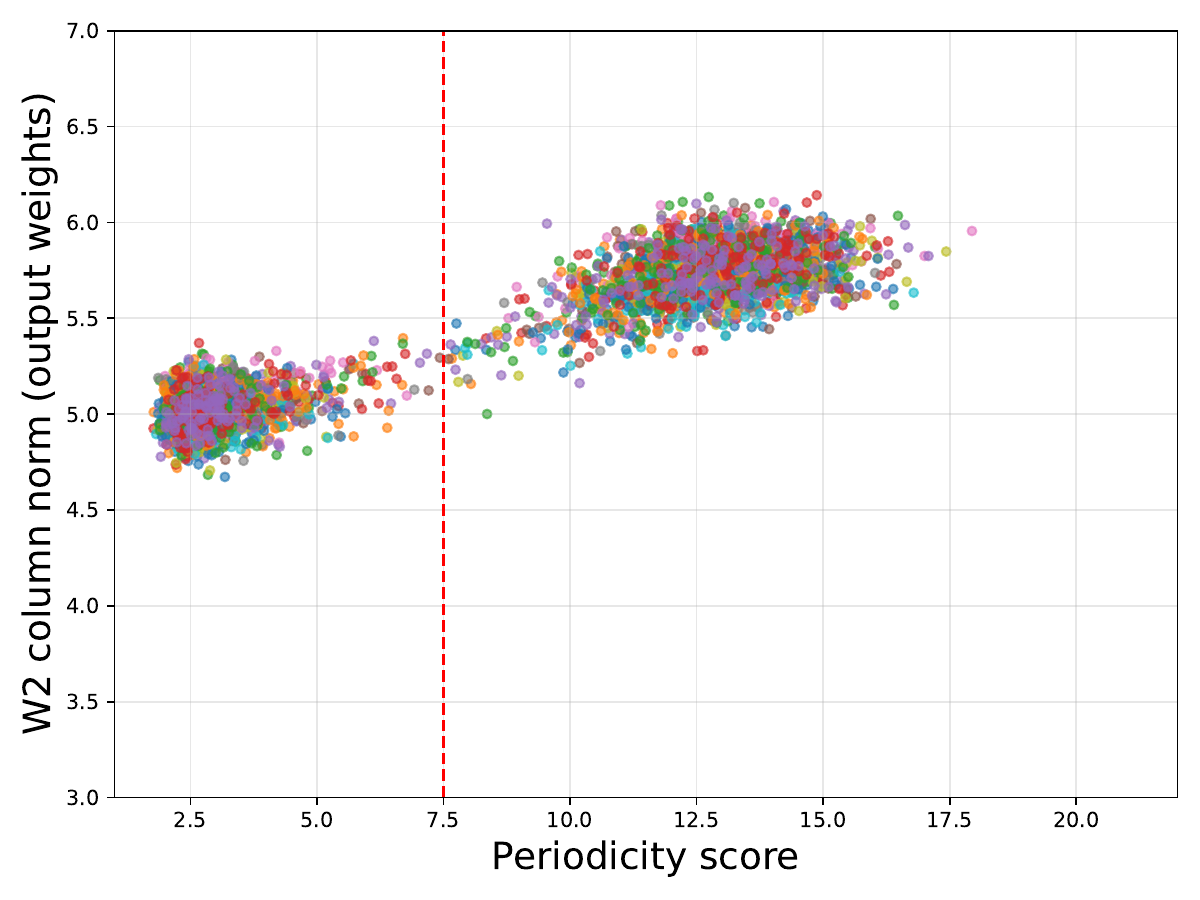}
        \caption{$\alpha = 0.10$}
    \end{subfigure}\hfill
    \begin{subfigure}[b]{0.32\linewidth}
        \includegraphics[width=\linewidth]{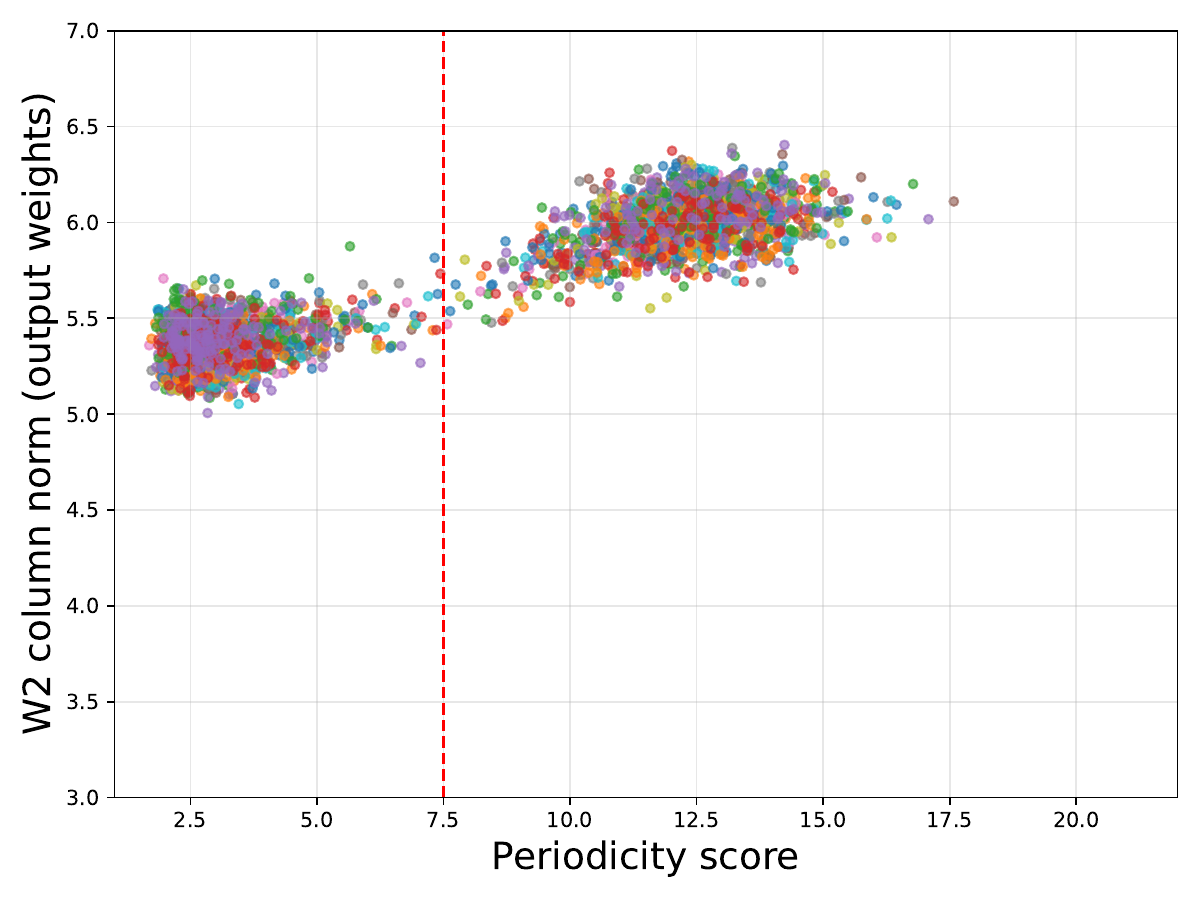}
        \caption{$\alpha = 0.15$}
    \end{subfigure}

    \vspace{0.5em}

    \begin{subfigure}[b]{0.32\linewidth}
        \includegraphics[width=\linewidth]{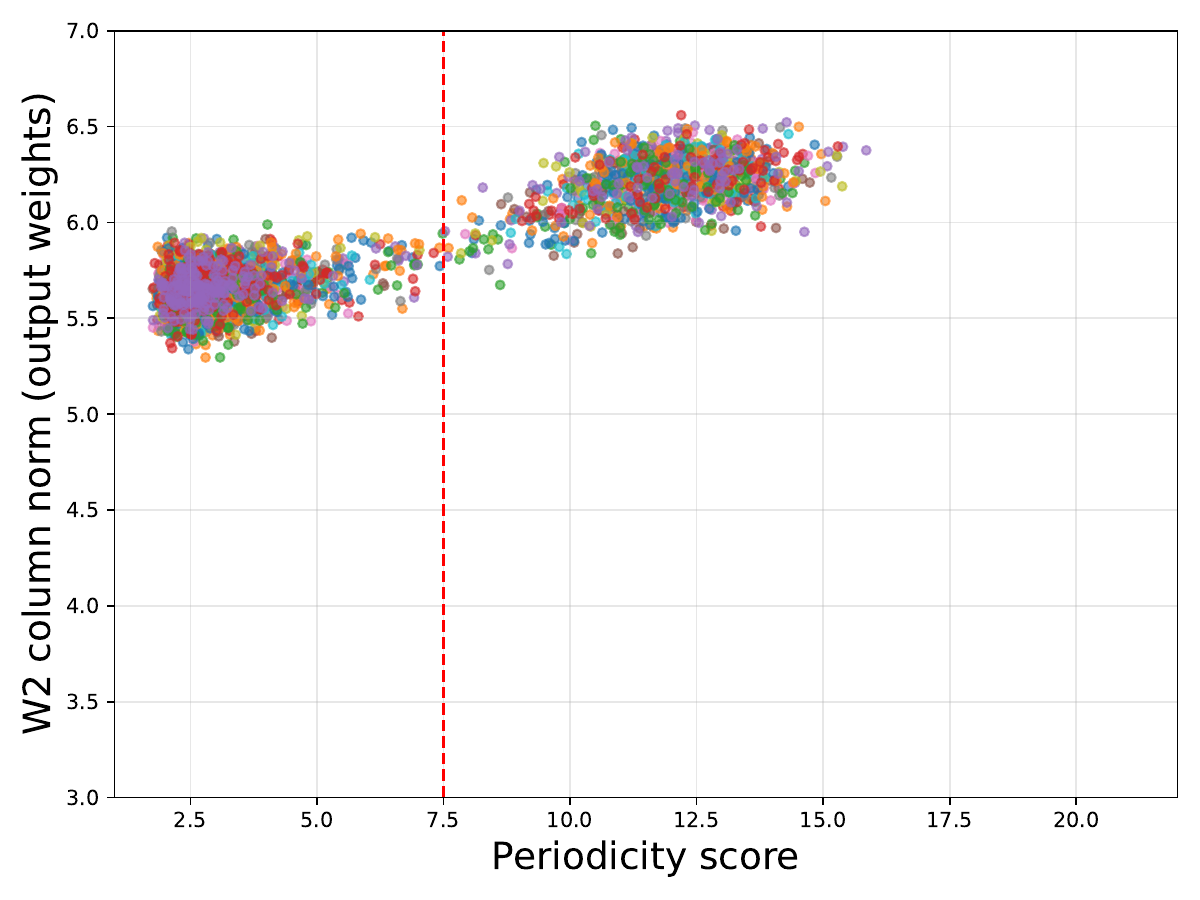}
        \caption{$\alpha = 0.20$}
    \end{subfigure}\hfill
    \begin{subfigure}[b]{0.32\linewidth}
        \includegraphics[width=\linewidth]{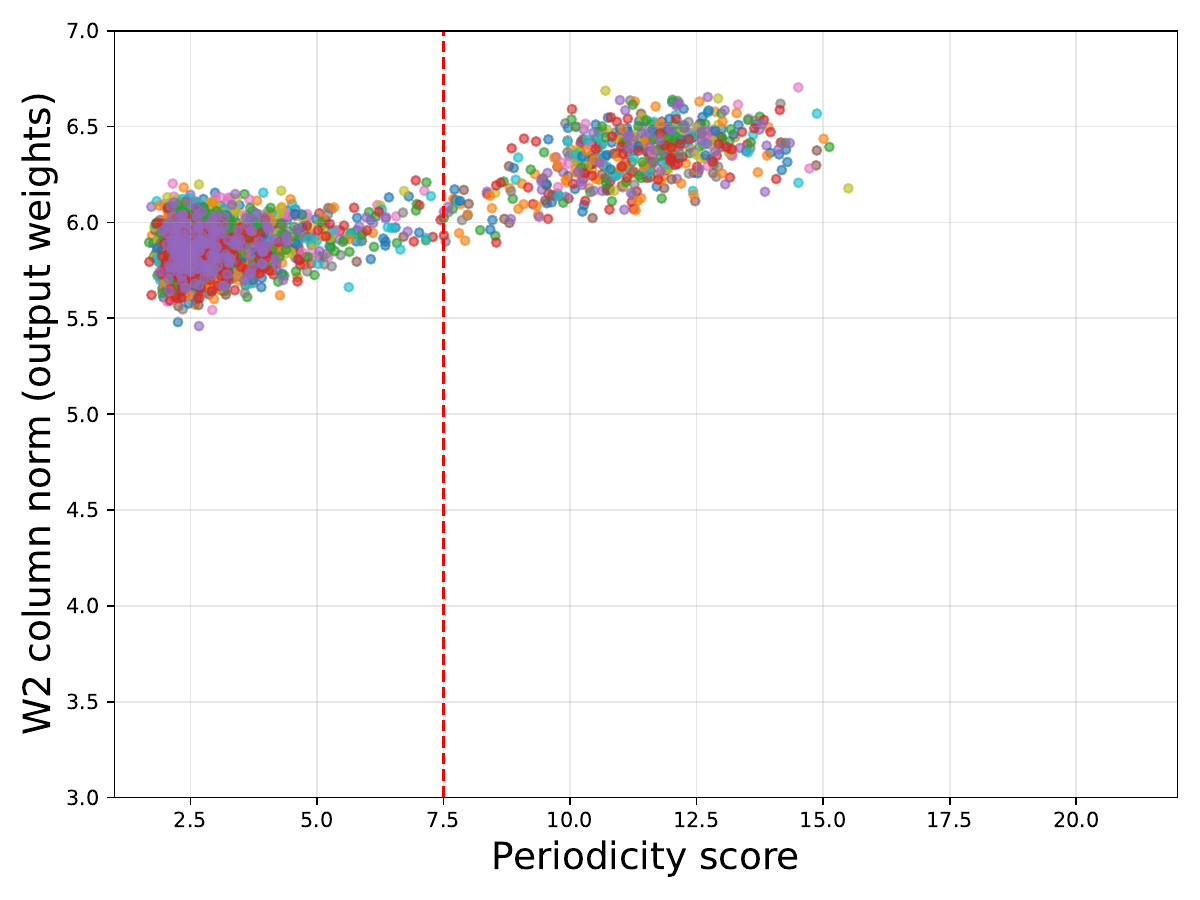}
        \caption{$\alpha = 0.25$}
    \end{subfigure}\hfill
    \begin{subfigure}[b]{0.32\linewidth}
        \includegraphics[width=\linewidth]{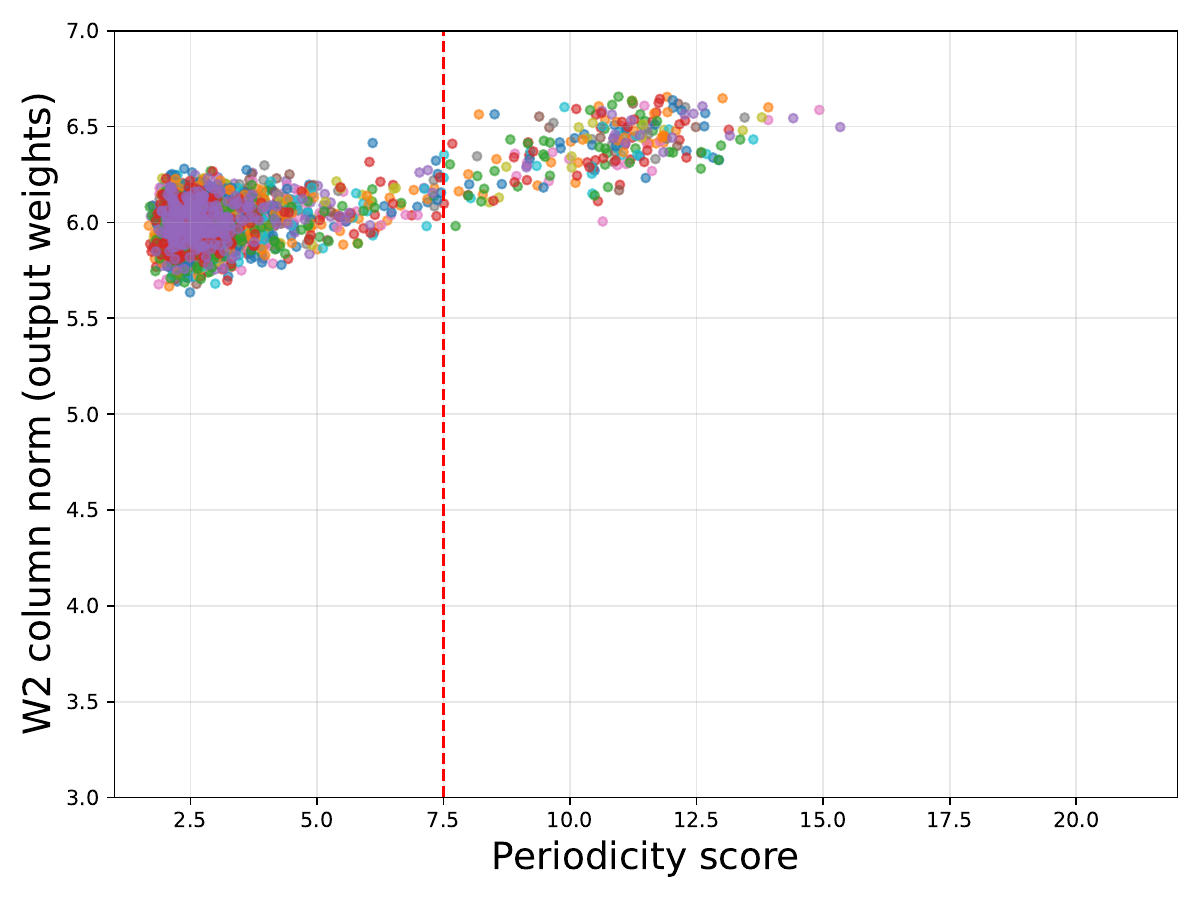}
        \caption{$\alpha = 0.30$}
    \end{subfigure}
    \caption{Periodicity vs.\ $L^2$ output weight norm across noise
        levels $\alpha$.}
    \label{fig:periodicity_alpha_grid}
\end{figure}

This gap becomes less pronounced at higher $\alpha$ levels, as more neurons
with intermediate periodicity values are present and there is no longer a
clear gap between the two populations (see
\Cref{fig:periodicity_alpha_grid}). We still use the same threshold ($7.5$)
in this case as it lies roughly at the midpoint of the full distribution
across all $\alpha$ levels and ensures consistency with the $\alpha = 0$
case during analysis.

Across all $\alpha$ levels, the periodicity/output weight norm correlation
is present, further supporting the observation that high-periodicity neurons
have the greatest impact on the final answer.

\newpage

\section{Square wave and cosine wave fit metric}
\label{apx:distance}

\begin{lemma}
\label{lma:4pi}
    Given a square wave with with amplitude $A$ and a sinusoidal wave
    with the same phase, frequency, and vertical offset, setting the
    amplitude of the sinusoidal wave to $\frac{4}{\pi} A$ minimizes the
    mean squared deviation between the two waves over one full period.
\end{lemma}

\begin{proof}
    We assume continuous $x$ values rather than the discrete indices used in the actual test. With the number of indices $p = 97$
    , the discrete case is a reasonably accurate approximation of the
    continuous case.

    Without loss of generality, we fix the square wave $f(x) =
    \sign(\sin(x))$. Then $A = 1$, so the corresponding sinusoidal wave
    is $g(x) = k \sin(x)$, where $k$, the constant that minimizes the
    mean squared deviation between $f(x)$ and $g(x)$, is to be
    determined.

    The mean squared deviation is
    \[
        \frac{1}{2\pi} \int_0^{2\pi} (f(x) - g(x))^2\,dx = \frac{1}{2\pi} \left(\int_0^{\pi} (1 - k \sin(x))^2\,dx + \int_{\pi}^{2\pi} (-1 - k \sin(x))^2\,dx\right),
    \]
    which evaluates to $\frac{1}{2\pi} (\pi k^2 - 8k + 2\pi)$. This is
    minimized at $k = \frac{- (- 8)}{2 (\pi)} = \frac{4}{\pi}$.
\end{proof}

\begin{figure}[h]
    \centering
    \includegraphics[width=\linewidth]{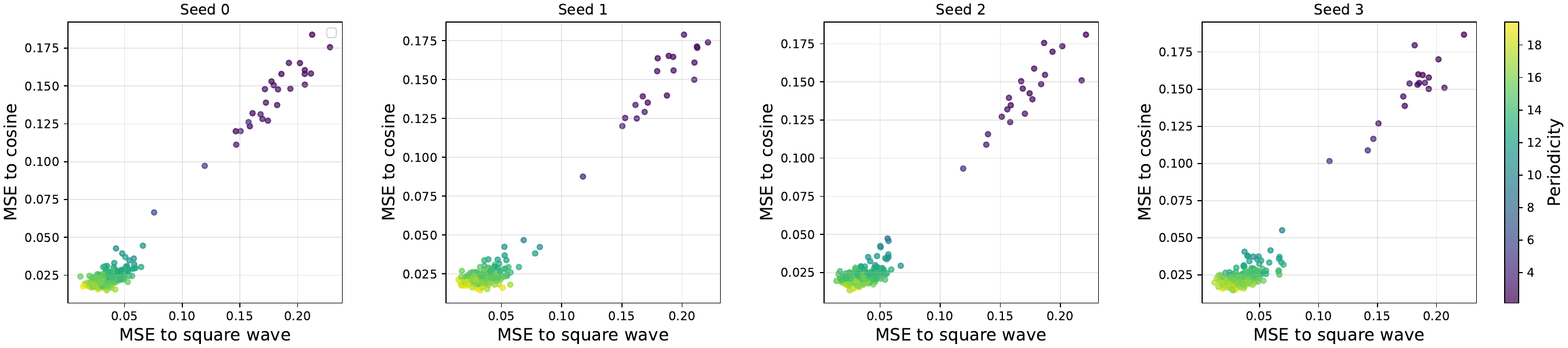}
    \caption{Mean squared deviation from the ideal square and cosine
    curves, for all 256 neurons across four seeds. Two clusters,
    representing structured and unstructured neurons, are clearly visible.}
    \label{fig:distance_scatter}
\end{figure}

We calculate the MSD between actual neuron weights against the ideal square
wave as well as the ideal cosine wave. The differences between the square
MSD and cosine MSD per neuron are plotted in \Cref{fig:distance_scatter}.

In order to control for bias in the distance measure, we perform the same
procedure on the unstructured neurons, pooled across all seeds. We find that
with these neurons, the MSD difference is $0.033$ on average.

We compare the distribution of the MSD difference between structured and
unstructured neurons. A one‑tailed Welch two‑sample $t$-test is performed,
with the null hypothesis that the mean difference for structured neurons
equals that of unstructured neurons, and the alternative hypothesis that the
mean difference is smaller for structured neurons (indicating that
structured neurons are, relative to the unstructured neurons, closer to a
square wave than to a cosine wave). The results across 15 seeds are
summarized in \Cref{tab:distance_ttest}. We find a highly significant
difference in every seed, confirming that the weights of structured neurons
fit a square wave more closely than a cosine wave.

\begin{table}[H]
\centering
\caption{Two-sample t-test results (MSE square minus MSE cosine, structured against unstructured neurons) across 15 seeds.}
\label{tab:distance_ttest}
\begin{tabular}{ccc}
\toprule
Seed & $t$ & $p$ \\
\midrule
0 & -10.571 & $<10^{-10}$ \\
1 & -9.027 & $<10^{-6}$ \\
2 & -6.540 & $<10^{-6}$ \\
3 & -6.247 & $<10^{-5}$ \\
4 & -6.410 & $<10^{-5}$ \\
5 & -6.845 & $<10^{-5}$ \\
6 & -7.977 & $<10^{-6}$ \\
7 & -9.703 & $<10^{-6}$ \\
8 & -9.851 & $<10^{-10}$ \\
9 & -9.287 & $<10^{-6}$ \\
10 & -10.635 & $<10^{-6}$ \\
11 & -5.966 & $<10^{-5}$ \\
12 & -7.546 & $<10^{-6}$ \\
13 & -7.550 & $<10^{-6}$ \\
14 & -5.457 & $<10^{-5}$ \\
\bottomrule
\end{tabular}
\end{table}

\newpage

\section{Extraction With Learned Amplitudes and Biases}
\label{apx:extraction_amp}

We extract the amplitudes and vertical offsets for the waves by taking $H$
as the 90-th percentile weight out of all positive weights for a given
$W_a\ti$, $W_b\ti$, or $W_{\out}\ti$, and likewise $L$ as the 10-th
percentile weight out of all negative weights. This is done to determine the
values that the square wave in the model's weights takes on in the positive
half-period and negative half-period respectively, ignoring outliers. We
then set the amplitude as $\frac{H - L}{2}$ and vertical offset as $\frac{H
+ L}{2}$, so the constructed square wave takes the value of $H$ in its
positive half-period and $L$ in its negative half-period.

In addition, we add each neuron's learned bias to the output of its
corresponding extracted neuron.

The accuracy achieved by this modified extraction method is shown in
\Cref{tab:extraction_amp_final,tab:extraction_amp_saturation}.

\begin{table}[h]
\centering
\caption{Final checkpoint validation accuracy with extracted amplitude and bias. Mean $\pm$ std over 15 seeds.}
\label{tab:extraction_amp_final}
\begin{tabular}{c|cccc}
\toprule
$\alpha$ & Real & \coscos & \sqcos & \sqsq \\
\midrule
0.00 & $0.9998\pm0.0001$ & $\mathbf{0.9997\pm0.0004}$ & $0.9991\pm0.0005$ & $0.9794\pm0.0067$ \\
0.05 & $0.9270\pm0.0139$ & $\mathbf{0.9979\pm0.0011}$ & $0.9964\pm0.0014$ & $0.9630\pm0.0076$ \\
0.10 & $0.5518\pm0.0334$ & $\mathbf{0.9928\pm0.0057}$ & $0.9926\pm0.0029$ & $0.9489\pm0.0132$ \\
0.15 & $0.2070\pm0.0273$ & $\mathbf{0.9906\pm0.0031}$ & $0.9897\pm0.0030$ & $0.9425\pm0.0063$ \\
0.20 & $0.0523\pm0.0093$ & $0.9791\pm0.0072$ & $\mathbf{0.9815\pm0.0048}$ & $0.9219\pm0.0140$ \\
0.25 & $0.0120\pm0.0027$ & $0.9621\pm0.0110$ & $\mathbf{0.9696\pm0.0070}$ & $0.8990\pm0.0169$ \\
0.30 & $0.0040\pm0.0011$ & $0.9360\pm0.0183$ & $\mathbf{0.9509\pm0.0113}$ & $0.8738\pm0.0261$ \\
\bottomrule
\end{tabular}
\end{table}
\begin{table}[h]
\centering
\caption{Saturation checkpoint validation accuracy with extracted amplitude and bias. Mean $\pm$ std over 15 seeds.}
\label{tab:extraction_amp_saturation}
\begin{tabular}{c|cccc}
\toprule
$\alpha$ & Real & \coscos & \sqcos & \sqsq \\
\midrule
0.00 & $0.0000\pm0.0000$ & $0.4482\pm0.0809$ & $\mathbf{0.5519\pm0.0717}$ & $0.4324\pm0.0621$ \\
0.05 & $0.0000\pm0.0000$ & $0.4622\pm0.0964$ & $\mathbf{0.5609\pm0.0879}$ & $0.4453\pm0.0781$ \\
0.10 & $0.0001\pm0.0001$ & $0.5188\pm0.0945$ & $\mathbf{0.6096\pm0.0905}$ & $0.4969\pm0.0831$ \\
0.15 & $0.0002\pm0.0001$ & $0.5472\pm0.0570$ & $\mathbf{0.6302\pm0.0468}$ & $0.4974\pm0.0483$ \\
0.20 & $0.0004\pm0.0002$ & $0.4855\pm0.0831$ & $\mathbf{0.5695\pm0.0789}$ & $0.4610\pm0.0715$ \\
0.25 & $0.0004\pm0.0003$ & $0.4333\pm0.0656$ & $\mathbf{0.5146\pm0.0628}$ & $0.4054\pm0.0602$ \\
0.30 & $0.0010\pm0.0003$ & $0.3926\pm0.0579$ & $\mathbf{0.4648\pm0.0604}$ & $0.3708\pm0.0473$ \\
\bottomrule
\end{tabular}
\end{table}

\newpage

\section{Extended Activation Ablation Results}
\label{apx:activation_full}

This appendix provides the complete set of metrics and additional visualizations
for the activation function ablation experiment described in
\Cref{sec:activation_ablation}. All results are derived from the same models
(three seeds per activation, final and saturation checkpoints) using the
periodicity threshold $>7.5$ to define structured neurons.

\subsection{Full Metric Tables}

\Cref{tab:activation_final_full,tab:activation_saturation_full} report, for
the final and saturation checkpoints respectively, the number of structured
neurons, mean periodicity score across all neurons, the mean bimodality,
frequency‑matching fractions (the fraction of all neurons where the dominant
FFT frequency in $W_a\ti$, $W_b\ti$, and $W_{\out}\ti$ were equal), circular
correlations between $\phi_a + \phi_b$ and $\phi_{\out}$ for all neurons
($r_{\mathrm{all}}$) and structured neurons only ($r_{\mathrm{high}}$),
extraction accuracies (\sqsq, \sqcos, and \coscos), validation accuracy,
structured-neurons‑only accuracy, and the fraction of dead neurons (that is,
neurons with all weights having absolute value $<10^{-4}$). Values are mean
$\pm$ standard deviation over three random seeds.

\begin{table}[h]
\scriptsize
\centering
\caption{Full activation ablation metrics, final checkpoint.}
\label{tab:activation_final_full}
\begin{tabular}{lcccccc}
\toprule

Metric & relu & gelu & tanh & silu & quadratic & quadratic\_nonzero \\
\midrule

$N_{\text{struct}}$ & 233.3$\pm$3.8 & 228.3$\pm$4.9 & 251.3$\pm$2.5 & 241.7$\pm$1.2 & 151.3$\pm$5.0 & 81.0$\pm$2.0 \\
Periodicity & 14.44$\pm$0.16 & 16.97$\pm$0.29 & 17.00$\pm$0.13 & 17.35$\pm$0.01 & 11.17$\pm$0.38 & 14.19$\pm$0.32 \\
Bimodality & 0.68$\pm$0.01 & 0.38$\pm$0.00 & 0.43$\pm$0.00 & 0.40$\pm$0.01 & 0.32$\pm$0.16 & 0.27$\pm$0.00 \\
Frequency Match & 0.95$\pm$0.01 & 0.92$\pm$0.01 & 0.99$\pm$0.01 & 0.96$\pm$0.00 & 0.71$\pm$0.02 & 0.70$\pm$0.03 \\
$r_{\text{all}}$ & 0.99$\pm$0.01 & 0.98$\pm$0.01 & 0.07$\pm$0.08 & 0.99$\pm$0.00 & 0.90$\pm$0.05 & 0.94$\pm$0.02 \\
$r_{\text{high}}$ & 1.00$\pm$0.00 & 1.00$\pm$0.00 & 0.05$\pm$0.07 & 1.00$\pm$0.00 & 0.94$\pm$0.09 & 1.00$\pm$0.00 \\
\sqsq & 0.983$\pm$0.003 & 0.962$\pm$0.007 & 0.009$\pm$0.001 & 0.935$\pm$0.017 & 0.996$\pm$0.005 & 0.996$\pm$0.005 \\
\sqcos & 0.999$\pm$0.000 & 0.997$\pm$0.001 & 0.008$\pm$0.001 & 0.995$\pm$0.001 & 1.000$\pm$0.000 & 1.000$\pm$0.000 \\
\coscos & 1.000$\pm$0.000 & 1.000$\pm$0.000 & 0.008$\pm$0.001 & 1.000$\pm$0.000 & 1.000$\pm$0.000 & 1.000$\pm$0.000 \\
Val Acc & 1.000$\pm$0.000 & 0.999$\pm$0.000 & 0.996$\pm$0.002 & 0.994$\pm$0.000 & 0.996$\pm$0.002 & 0.996$\pm$0.002 \\
Struct Acc & 0.999$\pm$0.000 & 0.984$\pm$0.004 & 0.996$\pm$0.001 & 0.956$\pm$0.011 & 0.985$\pm$0.005 & 0.985$\pm$0.005 \\
Dead Fraction & 0.00$\pm$0.00 & 0.00$\pm$0.00 & 0.00$\pm$0.00 & 0.00$\pm$0.00 & 0.47$\pm$0.01 & 0.00$\pm$0.00 \\
\bottomrule

\end{tabular}
\end{table}

\begin{table}[h]
\scriptsize
\centering
\caption{Full activation ablation metrics, saturation checkpoint.}
\label{tab:activation_saturation_full}
\begin{tabular}{lccccc}
\toprule

Metric & relu & gelu & tanh & silu & quadratic \\
\midrule

$N_{\text{struct}}$ & 0.0$\pm$0.0 & 0.0$\pm$0.0 & 0.0$\pm$0.0 & 0.0$\pm$0.0 & 0.3$\pm$0.6 \\
Periodicity & 2.57$\pm$0.03 & 2.46$\pm$0.03 & 2.44$\pm$0.04 & 2.48$\pm$0.01 & 2.82$\pm$0.08 \\
Bimodality & 0.38$\pm$0.01 & 0.36$\pm$0.00 & 0.25$\pm$0.00 & 0.31$\pm$0.00 & 0.49$\pm$0.01 \\
Frequency Match & 0.16$\pm$0.01 & 0.08$\pm$0.02 & 0.08$\pm$0.00 & 0.10$\pm$0.02 & 0.31$\pm$0.06 \\
$r_{\text{all}}$ & 0.68$\pm$0.12 & 0.83$\pm$0.04 & 0.01$\pm$0.10 & 0.85$\pm$0.01 & 0.86$\pm$0.06 \\
$r_{\text{high}}$ & --- & --- & --- & --- & --- \\
\sqsq & 0.734$\pm$0.041 & 0.759$\pm$0.020 & 0.010$\pm$0.001 & 0.670$\pm$0.022 & 0.997$\pm$0.001 \\
\sqcos & 0.849$\pm$0.032 & 0.884$\pm$0.013 & 0.010$\pm$0.001 & 0.785$\pm$0.023 & 1.000$\pm$0.000 \\
\coscos & 0.808$\pm$0.038 & 0.913$\pm$0.017 & 0.009$\pm$0.001 & 0.772$\pm$0.039 & 1.000$\pm$0.000 \\
Val Acc & 0.000$\pm$0.000 & 0.000$\pm$0.000 & 0.000$\pm$0.000 & 0.000$\pm$0.000 & 0.000$\pm$0.000 \\
Struct Acc & --- & --- & --- & --- & --- \\
Dead Fraction & 0.00$\pm$0.00 & 0.00$\pm$0.00 & 0.00$\pm$0.00 & 0.00$\pm$0.00 & 0.00$\pm$0.00 \\
\bottomrule

\end{tabular}
\end{table}

For the quadratic activation, a large fraction ($\sim 47\%$) of neurons are
dead. The column \texttt{quadratic\_nonzero} in
\Cref{tab:activation_final_full} shows the same metrics computed after
excluding these dead neurons; the remaining structured population behaves
similarly to the other activations, confirming that the dead neurons do not
participate in the learned algorithm.

\subsection{Harmonic Analysis}

\begin{figure}
    \begin{subfigure}{0.48\textwidth}
        \centering
        \includegraphics[width=\linewidth]{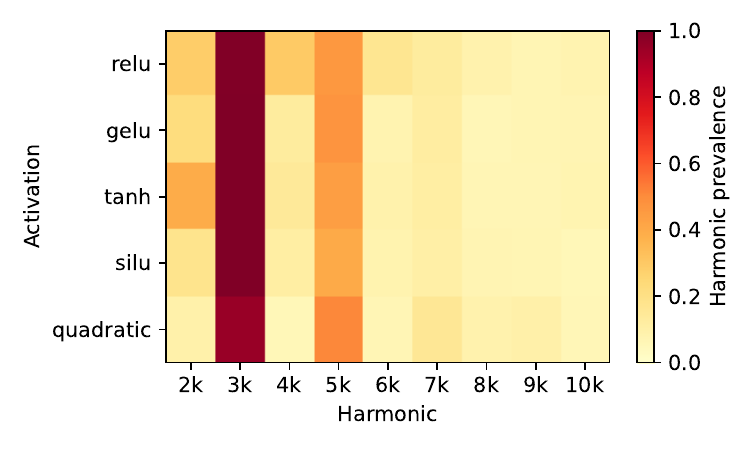}
        \caption{Prevalence}
        \label{fig:activation_harmonic_prevalence}
    \end{subfigure}
    \begin{subfigure}{0.48\textwidth}
        \centering
        \includegraphics[width=\linewidth]{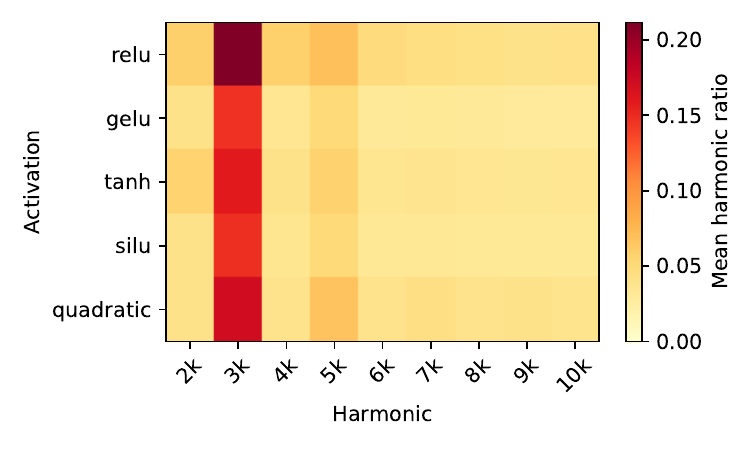}
        \caption{Ratio}
        \label{fig:activation_harmonic_ratio}
    \end{subfigure}
    \caption{Mean harmonic ratio and prevalence for structured neurons
    (final checkpoint). Activations on the y‑axis, harmonic frequency on
the x‑axis.}
\end{figure}

We define the \textit{prevalence} of a certain harmonic as the fraction of
neurons where that harmonic appears in the top 5 highest-weight-magnitude
harmonics, and the \textit{mean ratio} as the ratio between the weight
magnitude of the harmonic and the weight magnitude of the fundamental
frequency. \Cref{fig:activation_harmonic_prevalence} displays the prevalence
of each harmonic from $2k$ to $10k$, averaged across all final-checkpoint
structured neurons across all three tested seeds.
\Cref{fig:activation_harmonic_ratio} displays the mean ratios. We find that
the mean ratio of the $3k$ harmonic is significantly higher in the ReLU
models than any other model, indicating that the ReLU model weights are more
similar to square waves than the other activation models' weights.

\subsection{Periodicity Distributions}

\begin{figure}[h]
    \centering
    \includegraphics[width=\linewidth]{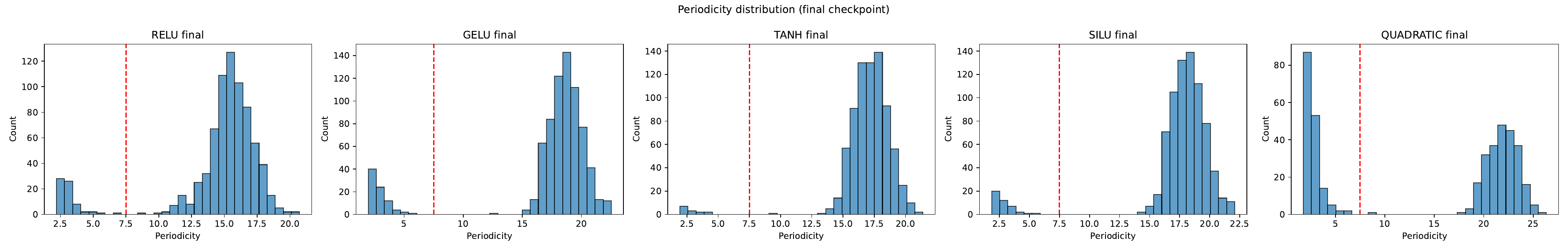}
    \caption{Periodicity score histograms for each activation (final checkpoint),
    combining all neurons across three seeds. The red dashed line indicates the
    threshold ($7.5$) separating structured and unstructured neurons.}
    \label{fig:periodicity_hists}
\end{figure}

\Cref{fig:periodicity_hists} shows the distribution of periodicity scores
for each activation at the final checkpoint, aggregated over three seeds. A
bimodal separation is visible for all activations. The vertical line marks
the threshold $7.5$ used to define structured neurons. This
threshold clearly splits the neurons into high- and low-periodicity groups
across all tested activations.

\subsection{Phase Scatter Grid}

\begin{figure}[h]
    \centering
    \includegraphics[width=\linewidth]{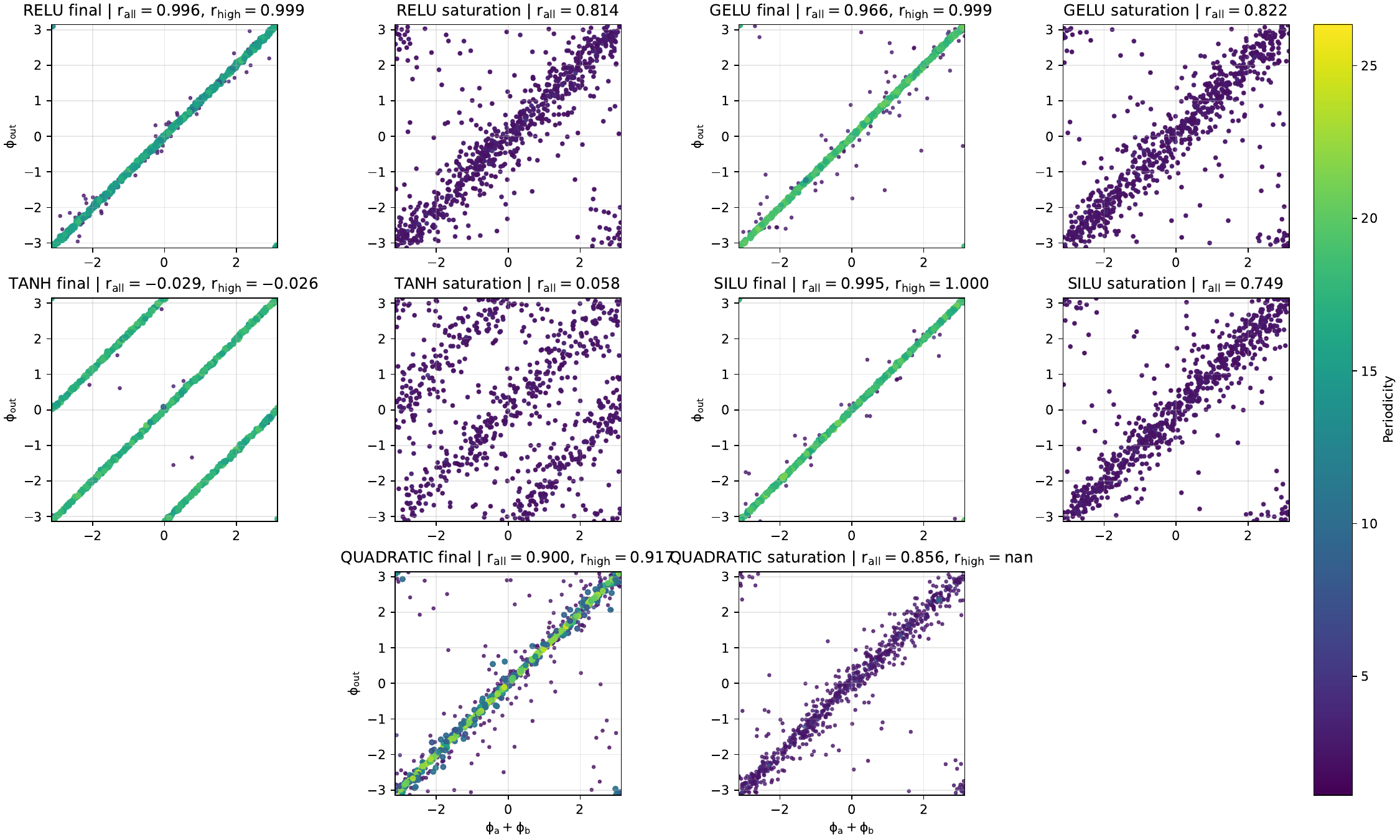}
    \caption{Phase scatter for all activations and checkpoints. Each panel shows
    $\phi_{\text{pred}} = \phi_a + \phi_b$ vs.\ $\phi_{\text{out}}$ wrapped to
    $[-\pi,\pi]$. Circular correlations $r_{\mathrm{all}}$ (across all neurons)
    and $r_{\mathrm{high}}$ (across only structured neurons) are shown.}
    \label{fig:phase_grid}
\end{figure}

\Cref{fig:phase_grid} displays the predicted phase $\phi_a + \phi_b$ against
the observed output phase $\phi_{\out}$ for every activation and both
checkpoints. Each panel aggregates all neurons across three seeds. The
diagonal alignment is nearly perfect for structured neurons across all
activations except Tanh, where the $\pi$ ambiguity produces two diagonals.
The saturation checkpoints (right column) show that structured neurons are
absent before grokking, yet a faint diagonal structure is already emerging.

\newpage

\section{Model Parameter Ablations}
\label{apx:param}

All mentioned metrics regarding input weights (periodicity, bimodality,
harmonics) are calculated from $W_a\ti$. Calculating the metrics on $W_b\ti$
yields identical qualitative results and near-identical qualitative results,
so we report only the results for $W_a\ti$ for brevity. All error bars in
graphs are $\pm 1$ standard deviation across 3 seeds.

\subsection{Hidden width}

\begin{figure}[h]
    \centering
    \begin{minipage}{0.48\textwidth}
        \centering
        \includegraphics[width=\linewidth]{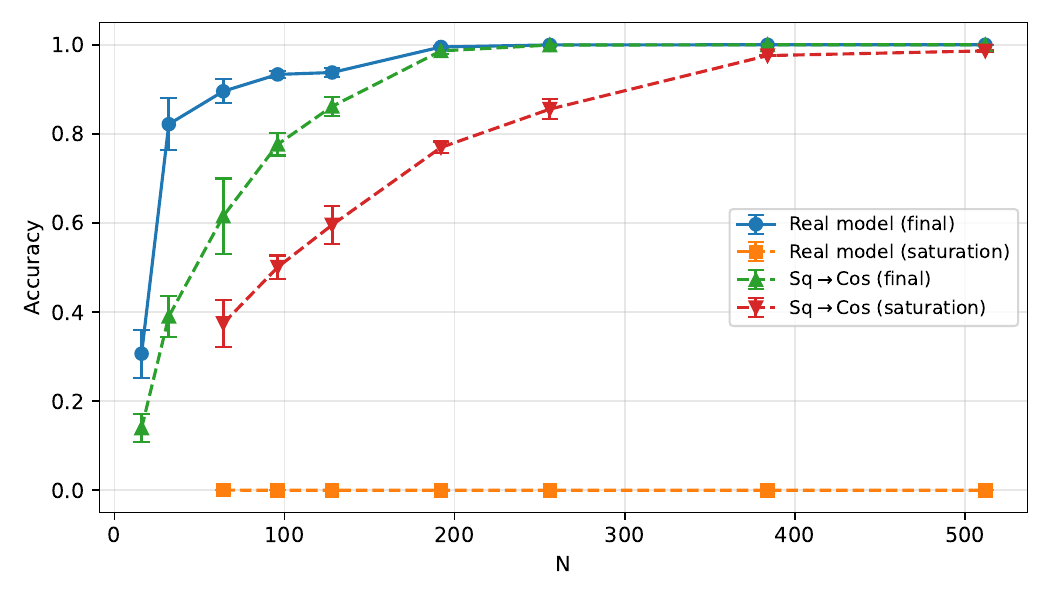}
        \caption{Real and extracted model accuracies across hidden widths.}
        \label{fig:accN}
    \end{minipage}
    \hfill
    \begin{minipage}{0.48\textwidth}
        \centering
        \includegraphics[width=\linewidth]{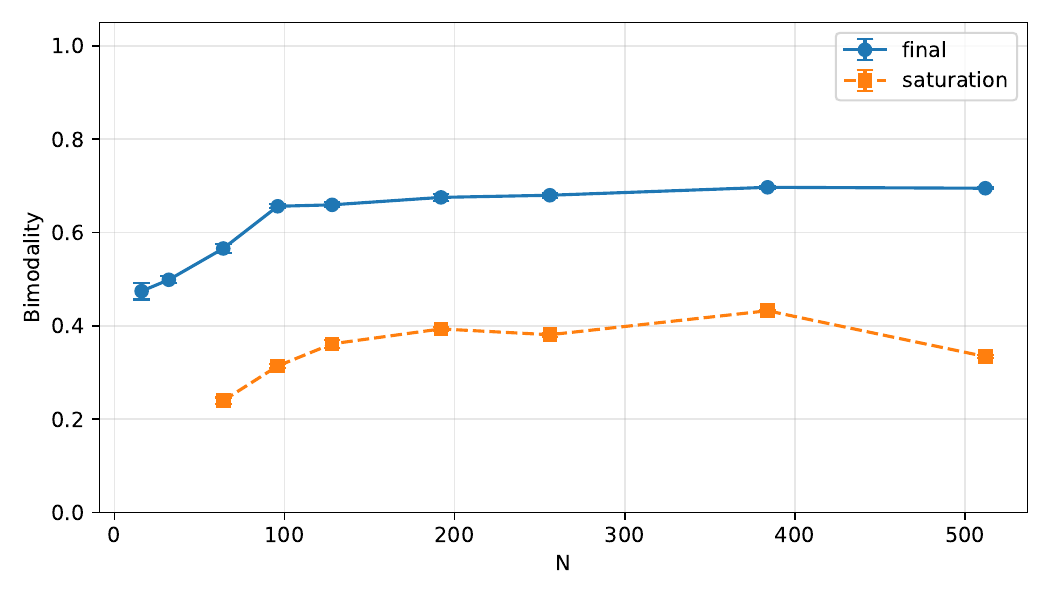}
        \caption{Weight bimodality across hidden widths.}
        \label{fig:bimodalityN}
    \end{minipage}
\end{figure}

\begin{figure}[h]
    \centering
    \includegraphics[width=\linewidth]{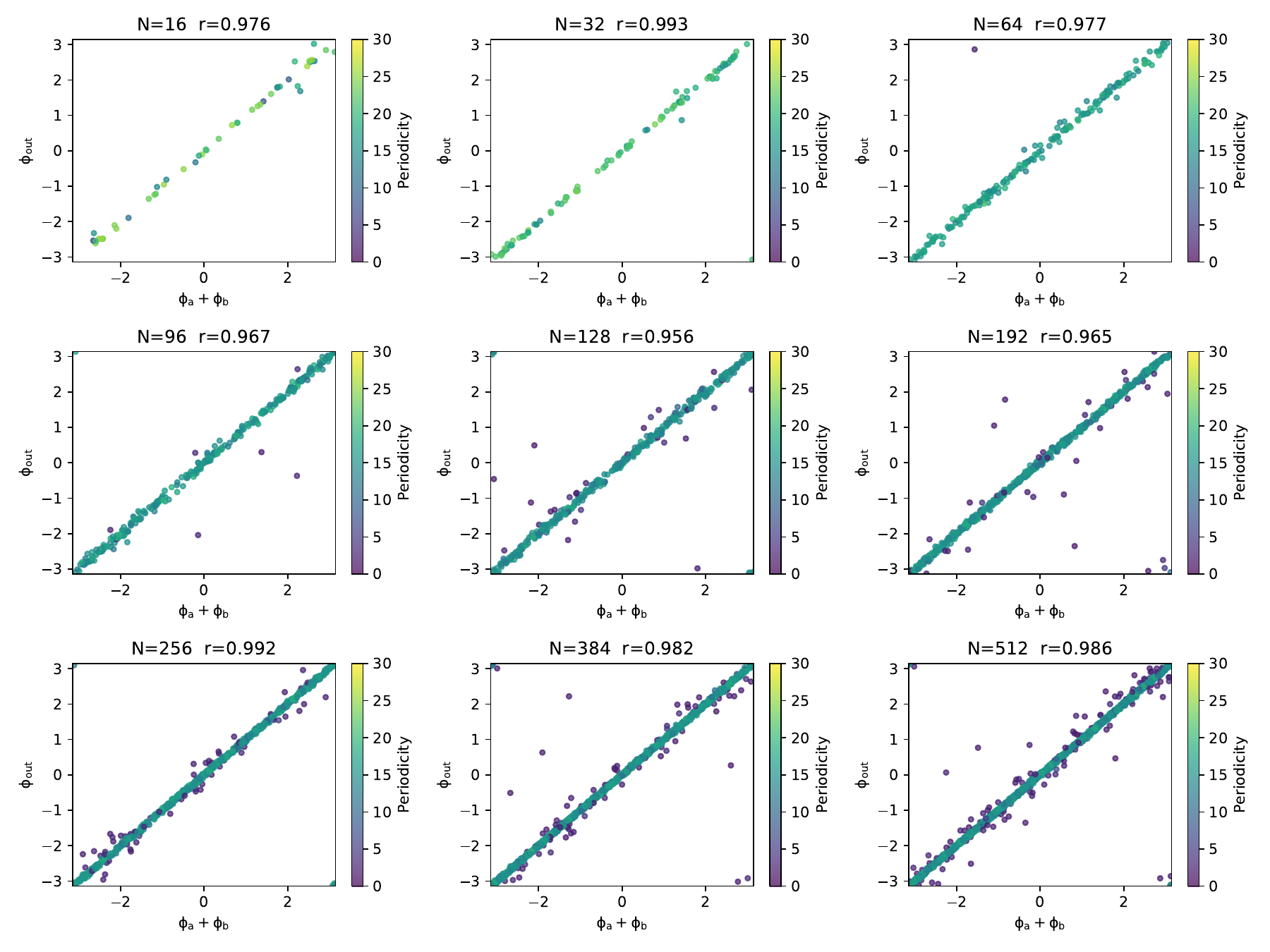}
    \caption{Predicted vs.\ actual $\phi_{\out}$ scatter and circular correlation across hidden widths.}
    \label{fig:phaseN}
\end{figure}

We test $N \in \{ 16, 32, 64, 96, 128, 192, 256, 384, 512 \}$. We find that
with $N \le 128$, the model is underparametrized and fails to achieve
near-perfect accuracy. With $N \ge 192$, the model achieves $>0.99$ accuracy
across all three tested seeds, as shown in \Cref{fig:accN}. Saturation
checkpoints consistently fail to achieve high accuracy. Extracted models
from the saturation checkpoints achieve much higher accuracy than the
saturation checkpoints themselves at all hidden widths, showing that latent
structure is encoded regardless of hidden width.

We find that the average bimodality across neurons in the final checkpoint
increases from roughly $0.48$ at $N = 16$ to roughly $0.65$ at $N = 96$, and
remaining close to $0.65$--$0.70$ across all $N \ge 128$, consistent with
the observation that all $N \ge 128$ exhibit square waves in input weights.
Bimodality is shown in \Cref{fig:bimodalityN}.

The phase-sum relation holds on all high-periodicity neurons in the final
checkpoint across all tested $N$, as shown in \Cref{fig:phaseN}. This
mirrors the presence of the phase-sum relation in all high-periodicity
neurons in the clean and noisy models in the main text.

Additionally, we measure the prevalence and mean ratio of harmonics across
structured neurons, in final checkpoints from each hidden width, with means
across all neurons and seeds shown in \Cref{fig:harmonicN}. We find that the
$3k$ and $5k$ harmonics increase in both magnitude and prevalence as $N$
increases, while other harmonics remain relatively low. This is consistent
with the odd-harmonic pattern of square waves; the $3k$ and $5k$ magnitudes
are, however, lower than the $\frac{1}{3}$ and $\frac{1}{5}$ expected from a
perfect square wave, and $7k$ and higher odd harmonics are not particularly
strong or prevalent. We hypothesize that this is due to the weights being
noisy rather than perfect square waves and the presence of intermediate
values at transitions.

\begin{figure}[h]
    \begin{subfigure}{0.48\textwidth}
        \centering
        \includegraphics[width=\linewidth]{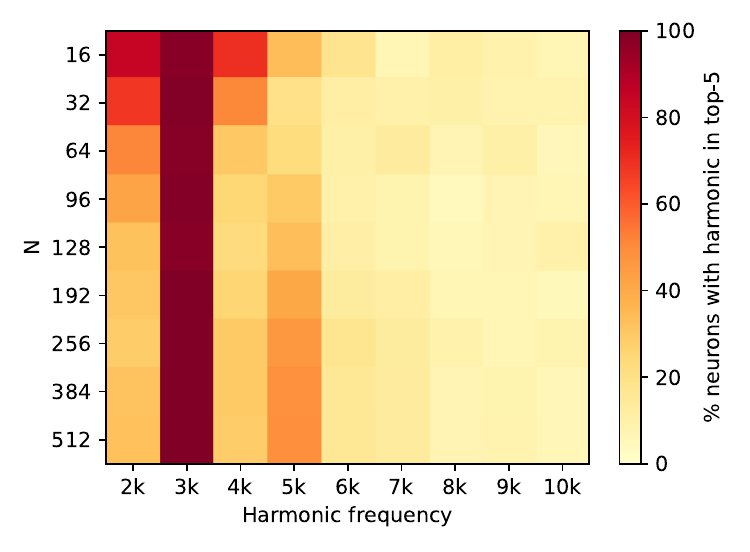}
        \caption{Prevalence}
        \label{fig:harmonicprevN}
    \end{subfigure}
    \begin{subfigure}{0.48\textwidth}
        \centering
        \includegraphics[width=\linewidth]{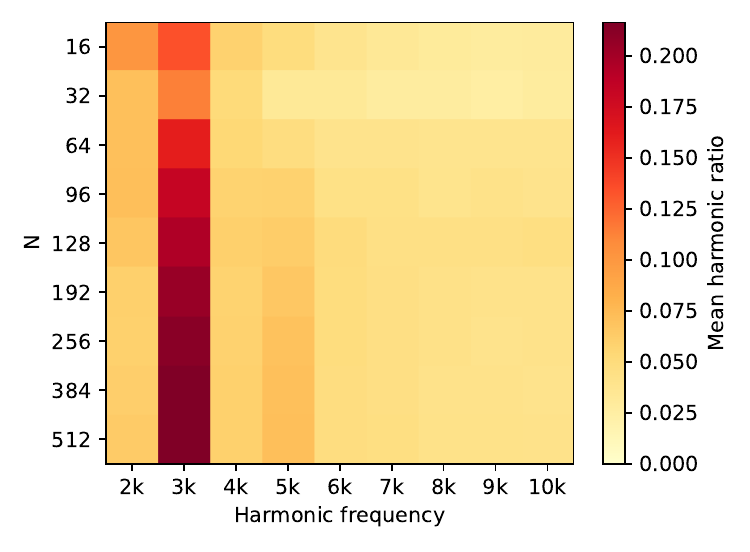}
        \caption{Ratio}
        \label{fig:harmonicratioN}
    \end{subfigure}
    \caption{Harmonic prevalence and ratio across hidden widths.}
    \label{fig:harmonicN}
\end{figure}

\subsection{Modulus}

\begin{figure}[h]
    \centering
    \begin{minipage}{0.48\textwidth}
        \centering
        \includegraphics[width=\linewidth]{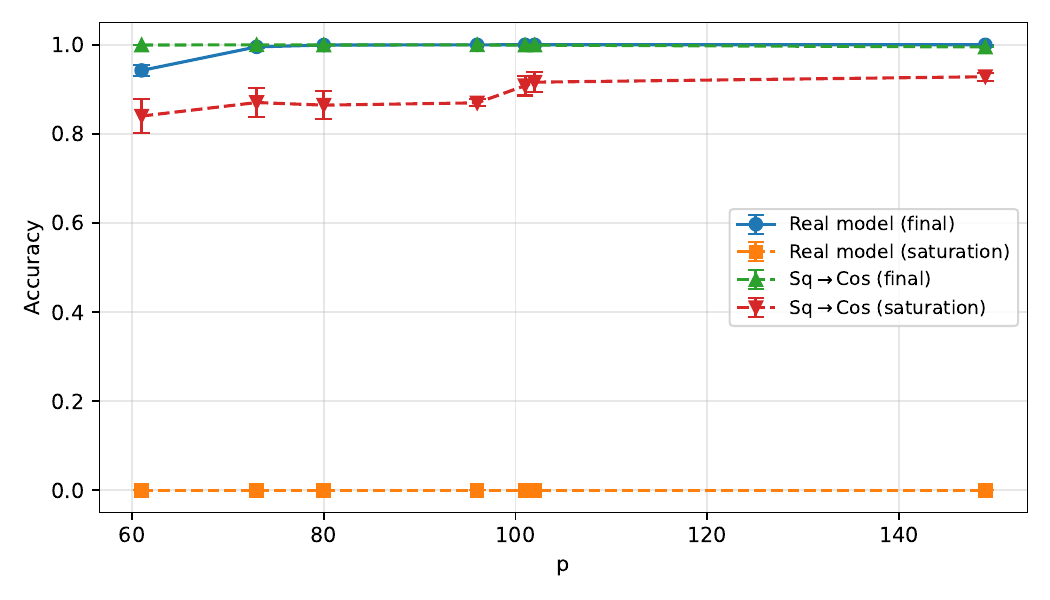}
        \caption{Real and extracted model accuracies across moduli.}
        \label{fig:accp}
    \end{minipage}
    \hfill
    \begin{minipage}{0.48\textwidth}
        \centering
        \includegraphics[width=\linewidth]{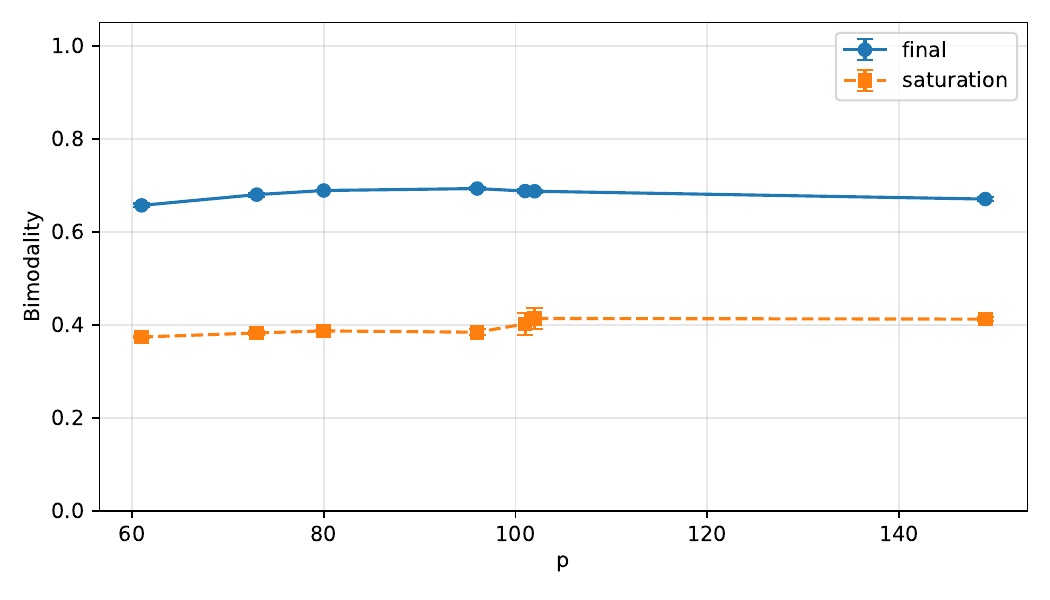}
        \caption{Weight bimodality across moduli.}
        \label{fig:bimodalityp}
    \end{minipage}
\end{figure}

\begin{figure}[h]
    \centering
    \includegraphics[width=\linewidth]{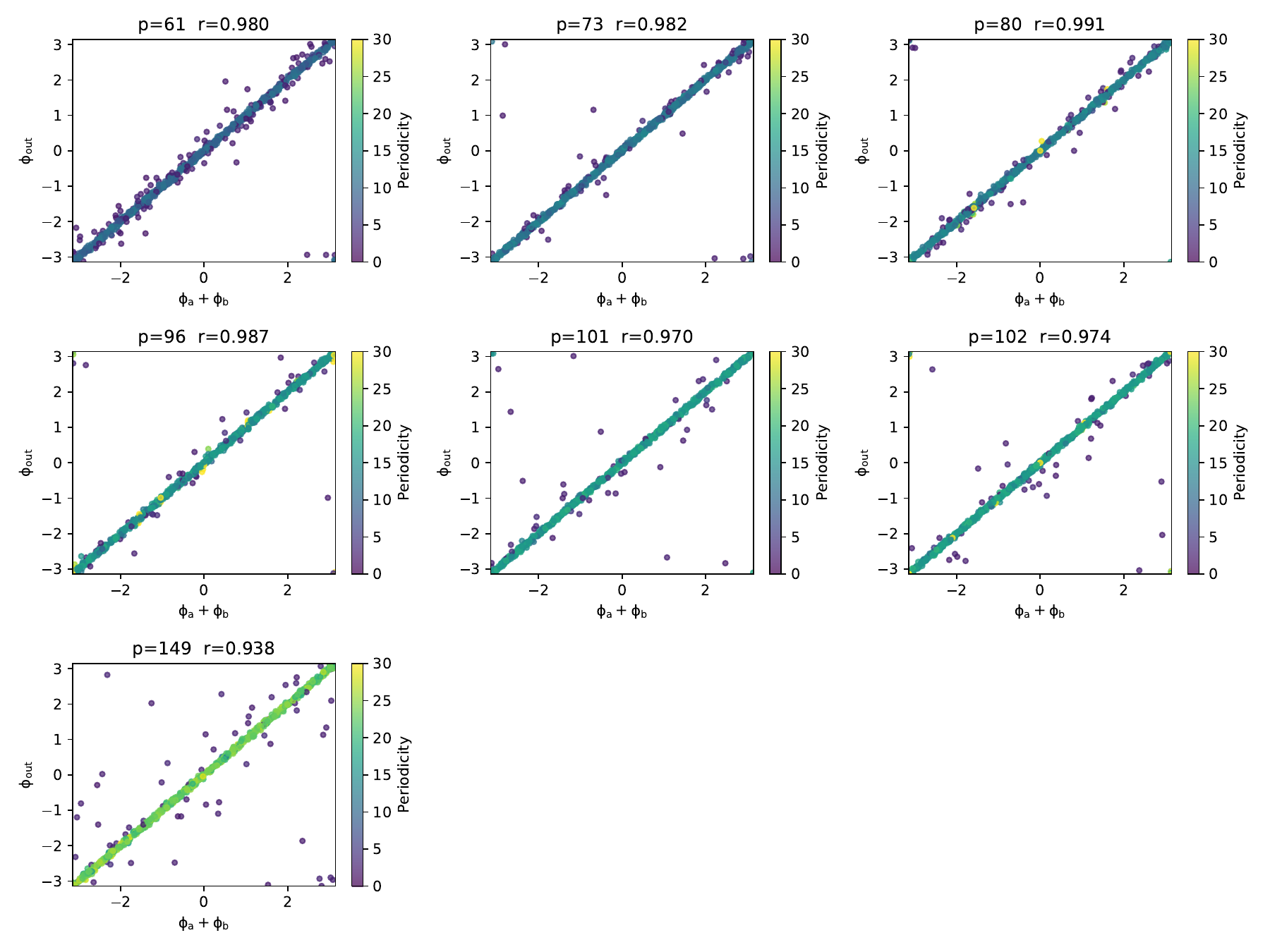}
    \caption{Predicted vs.\ actual $\phi_{\out}$ scatter and circular correlation across moduli.}
    \label{fig:phasep}
\end{figure}

We test $p \in \{ 61, 73, 80, 96, 101, 102, 149 \}$. We find that with $p =
61$, the model fails to reach near-perfect accuracy; we hypothesize that this is
due to the small size train set at this $p$ ($0.3 \cdot 61^2 \approx 1116$
training examples). With all other $p$, the model achieves $>0.99$ accuracy
across all three tested seeds, as shown in \Cref{fig:accp}. Saturation
checkpoints consistently fail to achieve high accuracy. As before, extracted
models from the saturation checkpoints achieve much higher accuracy than the
saturation checkpoints themselves with all tested $p$.

We find that the average bimodality across neurons in the final checkpoint
remains roughly equal across all choices of $p$, and likewise for average
bimodality across neurons in the saturation checkpoint. The final-checkpoint
result indicates that square waves emerge to the same degree across all $p$.
Bimodality is shown in \Cref{fig:bimodalityp}.

All tested $p$ result in models with a bimodal periodicity distribution,
with the phase-sum relation holding on all high-periodicity neurons in the
final checkpoint (\cref{fig:phasep}). This suggests the phase-sum relation
and algorithm for modular addition is preserved across moduli.

Prevalence and mean ratio of harmonics are shown in \Cref{fig:harmonicN}.
The $3k$ and $5k$ harmonics are consistently the highest-magnitude and most
prevalent harmonics.

\begin{figure}[h]
    \begin{subfigure}{0.48\textwidth}
        \centering
        \includegraphics[width=\linewidth]{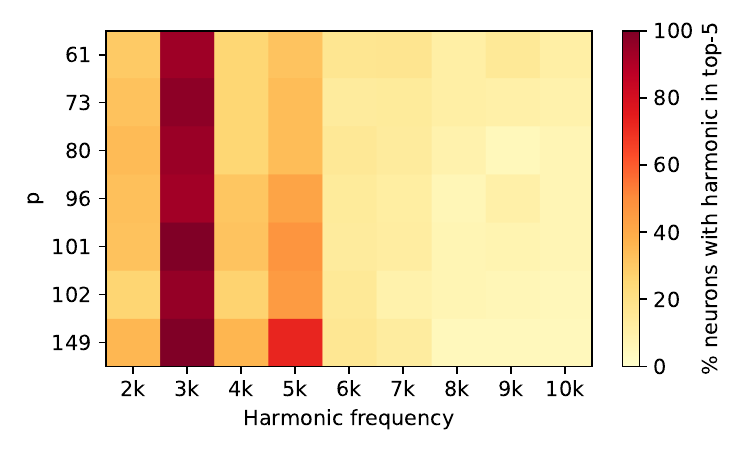}
        \caption{Prevalence}
        \label{fig:harmonicprevp}
    \end{subfigure}
    \begin{subfigure}{0.48\textwidth}
        \centering
        \includegraphics[width=\linewidth]{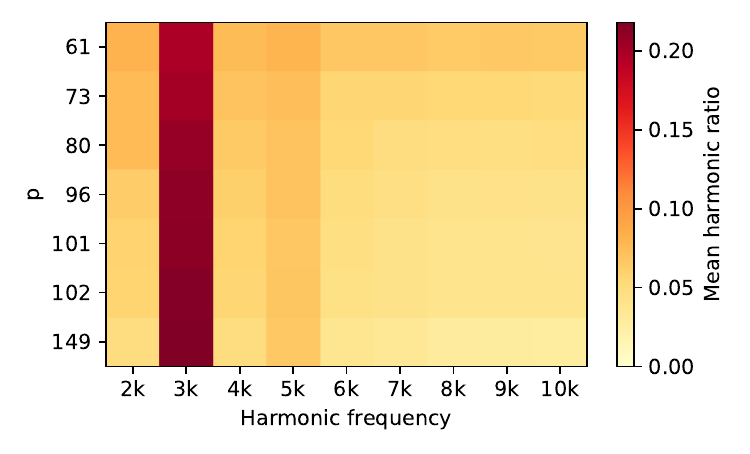}
        \caption{Ratio}
        \label{fig:harmonicratiop}
    \end{subfigure}
    \caption{Harmonic prevalence and ratio across moduli.}
    \label{fig:harmonicN}
\end{figure}

\newpage

\section{Compute}
\label{apx:compute}

All training can be completed in approximately 32 hours on an NVIDIA GeForce
RTX 3060 Mobile GPU, including all 15 seeds. All analyses (figures and
tables) can be run in under an hour on an RTX 3060 Mobile. The total
compute, including failed experiments, was approximately 72 hours.

\end{document}